\documentclass{article}

\usepackage[final, nonatbib]{neurips_2019}

\makeatletter
\renewcommand*{\@fnsymbol}[1]{\ensuremath{\ifcase#1\or \else\@ctrerr\fi}}
\makeatother

\usepackage[utf8]{inputenc} 
\usepackage[T1]{fontenc}    
\usepackage{hyperref}       
\usepackage{url}            
\usepackage{booktabs}       
\usepackage{amsfonts}       
\usepackage{nicefrac}       
\usepackage{microtype}      
\usepackage{amsmath}
\usepackage{amsthm}
\usepackage{amssymb}
\usepackage{physics}
\usepackage[ruled,vlined]{algorithm2e}
\usepackage{multirow}
\usepackage{graphicx}
\usepackage[belowskip=-6pt]{caption}
\usepackage{graphicx}
\usepackage{amsmath}
\usepackage{subcaption}

\usepackage{graphicx}
\usepackage{amssymb}
\usepackage{multirow}

\makeatletter

\usepackage{xcolor}

\usepackage{lineno}
\usepackage{amsmath}
\usepackage{subcaption}

\author{%
  Levi D. McClenny \ \ \ \ \ \ Ulisses Braga-Neto \thanks{$\!\!\!\!\!\!\!\!\!\!$Published in {\em Journal of Computational Physics}, Vol.\ 474, 111722.}\\ 
  Department of Electrical and Computer Engineering\\
  Texas A\&M University\\
  College Station, TX\ USA \\
  \texttt{\{levimcclenny,ulisses\}@tamu.edu} 
}

\title{Self-Adaptive Physics-Informed Neural Networks using a Soft Attention Mechanism}

\begin{document}

\maketitle

\begin{abstract}
Physics-Informed Neural Networks (PINNs) have emerged recently as a promising application of deep neural networks to the numerical solution of nonlinear partial differential equations (PDEs). However, it has been recognized that adaptive procedures are needed to force the neural network to fit accurately the stubborn spots in the solution of “stiff” PDEs. In this paper, we propose a fundamentally new way to train PINNs adaptively, where the adaptation weights are fully trainable and applied to each training point individually, so the neural network learns autonomously which regions of the solution are difficult and is forced to focus on them. The self-adaptation weights specify a soft multiplicative soft attention mask, which is reminiscent of similar mechanisms used in computer vision. The basic idea behind these SA-PINNs is to make the weights increase as the corresponding losses increase, which is accomplished by training the network to simultaneously minimize the losses and maximize the weights. In addition, we show how to build a continuous map of self-adaptive weights using Gaussian Process regression, which allows the use of stochastic gradient descent in problems where conventional gradient descent is not enough to produce accurate solutions. Finally, we derive the Neural Tangent Kernel matrix for SA-PINNs and use it to obtain a heuristic understanding of the effect of the self-adaptive weights on the dynamics of training in the limiting case of infinitely-wide PINNs, which suggests that SA-PINNs work by producing a smooth equalization of the eigenvalues of the NTK matrix corresponding to the different loss terms. In numerical experiments with several linear and nonlinear benchmark problems, the SA-PINN outperformed other state-of-the-art PINN algorithm in L2 error, while using a smaller number of training epochs.
\end{abstract}

\section{Introduction}

As part of the burgeoning field of scientific machine learning~\cite{osti_1478744}, physics-informed neural networks (PINNs) have emerged recently as an alternative to traditional numerical methods for partial different equations (PDE)~\cite{raissi2019physics,raissi2018forward,wight2020solving,wang2020and}. Typical data-driven deep learning methodologies do not take into account physical understanding of the problem domain. 
The PINN approach is based on a strong physics prior that constrains the output of a deep neural network by means of a system of PDEs. The potential of using neural networks as universal function approximators to solve PDEs had been recognized since the 1990's~\cite{dissanayake1994neural}. However, PINNs promise to take this approach to a different level by using deep neural networks, which is made possible by the vast advances in computational capabilities and training algorithms since that time~\cite{abadi2016tensorflow, revels2016forward}, as well as the availability of automatic differentiation methods~\cite{baydin2017automatic,paszke2017automatic}. 

A great advantage of PINNs over traditional time-stepping PDE solvers is that it is possible to obtain the solution over the entire spatial-temporal domain at once, using training points distributed irregularly across the domain, obviating the need of constructing computationally-expensive grids. In addition, the PINN solution defines a function over the continuous domain, rather that a discrete solution on a grid as in traditional methods. Finally, PINNs allow sample data assimilation in a natural and efficient way.

The continuous PINN algorithm proposed in \cite{raissi2019physics}, henceforth referred to as the ``baseline PINN'' algorithm,  
is effective at estimating solutions that are reasonably smooth with simple boundary conditions, such as the viscous Burgers,  Poisson and Schr\"{o}dinger PDEs. On the other hand, it has been observed that the baseline PINN has convergence and accuracy problems when solving ``stiff'' PDEs \cite{burden1985numerical}, with solutions that contain sharp space transitions or fast time evolution~\cite{wang2020understanding}. This is known to be the case, for example, when attempting to solve the nonlinear Allen-Cahn equation with the baseline PINN~\cite{wight2020solving}. As we will see in this paper, this may occur even in the case of the linear wave and advection PDEs.


This paper introduces Self-Adaptive PINNs (SA-PINNs), a fundamentally new method to train PINNs adaptively, which addresses the issues mentioned previously. SA-PINNs applies trainable weights on each training point, in a way that is reminiscent of soft multiplicative attention masks used in computer vision~\cite{wang2017residual,pang2019mask}.
The adaptation weights are trained concurrently with the network weights. As a result, initial, boundary or residue  points in difficult regions of the solution are automatically weighted more in the loss function, forcing the approximation to improve on those points. The basic principle in SA-PINNs is to make the weights increase as the corresponding losses do, which is accomplished by training the network to simultaneously minimize the losses and maximize the weights, i.e., to find a saddle point in the cost surface. 
 
We also propose a methodology to build a continuous map of self-adaptive weights based on Gaussian Process regression, in order to allow the use of stochastic gradient descent in training self-adaptive PINNs. This is illustrated by application to a 1-D wave PDE that is challenging to non-SGD training. 

Finally, we derive the Neural Tangent Kernel matrix for self-adaptive PINNs and use it to obtain a heuristic understanding of the effect of the self-adaptive weights on the dynamics of training in the limiting case of infinitely-wide PINNs. We examine the effect of the self-adaptive weights on the eigenvalues of the NTK matrix in the solution of a linear advection PDE, and observe that it not only equalizes the magnitudes between the different loss components, but also smooths the shape of the distribution of eigenvalues. This provides preliminary theoretical justification of the success of self-adaptive PINNs.

Comprehensive experimental results presented throughout the test, based on several well-known benchmarks, show that self-adaptive PINNs can solve ``stiff'' PDEs with significantly better accuracy than other state-of-the-art PINN algorithms, while using a smaller number of training epochs. 

\section{Background}
\subsection{Physics-Informed Neural Networks}

\def\bx{\boldsymbol{x}}
\def\bw{\boldsymbol{w}}

Consider the initial-boundary value problem:
\begin{align}
    & \mathcal{N}_{\bx,t}[ u(\bx, t)] \,=\, f(\bx,t)\,, \ \ \bx \in \Omega\,, \ t \in (0,T]\,, \label{eq:PDE} \\
    & \mathcal{B}_{\bx,t}[ u(\bx, t)] \,=\, g(\bx,t)\,, \ \  \bx \in \partial\Omega, \ t \in (0,T]\,,\label{eq:boundary} \\    
    & u(\bx, 0) = h(\bx)\,, \ \
    \bx \in \overline{\Omega}\,. \label{eq:initial}
\end{align}
Here, the domain $\Omega \subset R^d$ in a open set, $\overline{\Omega}$ is its closure, $u:\overline{\Omega}\times [0,T] \rightarrow R$ is the desired solution, $\boldsymbol{x} \in \Omega$ is a spatial vector variable, $t$ is time, and $\mathcal{N}_{\boldsymbol{x},t}$ and $\mathcal{B}_{\boldsymbol{x},t}$ are spatial-temporal differential operators. The problem {\em data} is provided by the forcing function $f:\Omega \rightarrow R$, the boundary condition function $g:\partial \Omega \times (0,T]$, and the initial condition function $h:\overline{\Omega} \rightarrow R$. Additionally, sensor data in the interior of the domain may be available. In any case, we assume that the data are sufficient and appropriate for a well-posed problem. Time-independent problems and other types of data can be handled similarly, so we will use the equations (1)-(3) as a model.

Following~\cite{raissi2019physics}, let $u(\bx, t)$ be approximated by the output $u(\bx,t;\bw)$ of a deep neural network with inputs $\bx$ and $t$ (in the case of a PDE system, this would be a neural network with multiple outputs). The value of $\mathcal{N}_{\bx,t}[ u(\bx,t;\bw)]$
and $\mathcal{B}_{\bx,t}[ u(\bx,t;\bw)]$ 
can be computed quickly and accurately using 
reverse-mode {\em automatic differentiation}~\cite{baydin2017automatic,paszke2017automatic}.

The network weights $\bw$ are trained by minimizing a loss function that penalizes the output for not satisfying (1)-(3): 
\begin{equation}
    \mathcal{L}(\bw) = \mathcal{L}_s(\bw) + \mathcal{L}_r(\bw) + \mathcal{L}_{b}(\bw) + \mathcal{L}_{0}(\bw)\,, 
\label{eq:loss}
\end{equation}
where $\mathcal{L}_s$ is the loss term corresponding to sample data (if any), while  $\mathcal{L}_r$, $\mathcal{L}_{b}$, and $\mathcal{L}_{0}$ are loss terms corresponding to not satisfying the PDE (\ref{eq:PDE}), the boundary condition (\ref{eq:boundary}), and the initial condition~(\ref{eq:initial}), respectively:
\begin{align}
    \mathcal{L}_s(\bw) &= \frac{1}{2}\sum^{N_s}_{i=1} |u(\bx^i_s, t^i_s;\bw) - y^i_s|^2, \\
    \mathcal{L}_r(\bw)&= \frac{1}{2}\sum^{N_r}_{i=1} |\mathcal{N}_{\bx,t}[u(\bx^i_r,t^i_r;\bw)]-f(\bx^i_r,t^i_r)|^2, \\
    \mathcal{L}_{b}(\bw) &= \frac{1}{2}\sum^{N_b}_{i=1}|\mathcal{B}_{\bx,t}[u(\bx^i_b, t^i_b;\bw)] - g(\bx^i_b,t^i_b)|^2, \, \label{Lb}\\
    \mathcal{L}_{0}(\bw) &= \frac{1}{2}\sum^{N_0}_{i=1}|u(\bx^i_0,0;\bw)- h(\bx^i_0)|^2.
\end{align}
where $\{\bx_s^i, t_s^i, y_s^i\}_{i = 1}^{N_s}$ are sensor data (if any), $\{\bx_0^i\}_{i = 1}^{N_0} $ are initial condition points,  $\{\bx_b^i, t^i_b\}_{i = 1}^{N_b}$ are boundary condition points, $\{\bx_r^i, t^i_r\}_{i = 1}^{N_r} $ are residue (``collocation'') points randomly distributed in the domain $\Omega$, and $N_s, N_0, N_b$ and $N_r$ denote the total number of sensor, initial, boundary, and residue points, respectively. 
The network weights $\bw$ can be tuned by minimizing the total training loss $\mathcal{L}(\bw)$ via standard gradient descent procedures used in deep learning.  

\subsection{Related Work}
\label{Sec:rw}

The baseline PINN algorithm described in the previous section, though remarkably successful in the solution of many linear and nonlinear PDEs, can produce inaccurate approximations, or fail to converge entirely, in the solution of certain ``stiff'' PDEs. A large amount of evidence has accumulated indicating that this happens due to the shortcomings of gradient descent applied to the multi-part or multi-objective loss function (\ref{eq:loss}); e.g., see \cite{wight2020solving,shin2020convergence,wang2020understanding,wang2020and}. This occurs because gradient descent is a greedy procedure that may latch on some of the components at the expense of others, which creates imbalance in the rate of descent among the different loss components and prevents convergence to the correct solution. The standard approach in the literature of PINNs to try to correct the imbalance is the introduction of weights in (\ref{eq:loss}):   
\begin{equation}
    \mathcal{L}(\bw) = \lambda_s\mathcal{L}_s(\bw) + \lambda_r\mathcal{L}_r(\bw) + \lambda_b \mathcal{L}_{b}(\bw) + \lambda_0\mathcal{L}_{0}(\bw)\,, 
\label{eq:loss_weighted}
\end{equation}
Several methods, from very simple to complex, have been advanced to set the values of these weights; we mention a few below. 

\paragraph{Nonadaptive Weighting} In \cite{wight2020solving}, it was pointed out that a premium should be put on forcing the neural network to satisfy the initial conditions closely, especially for PDEs describing time-irreversible processes, where the solution has to be approximated well early. Accordingly, a loss function of the form $\mathcal{L}(\theta) = \mathcal{L}_r(\theta) + \mathcal{L}_b(\theta) + C\,\mathcal{L}_0(\theta)$ was suggested, where $C \gg 1$ is a hyperparameter. 

\paragraph{Learning Rate Annealing} In~\cite{wang2020understanding}, it is argued that the optimal value of the weight $C$ in the previous scheme may vary wildly among different PDEs so that choosing its value would be difficult. Instead they propose to use weights that are tuned during training using statistics of the backpropagated gradients of the loss function. It is noteworthy that the weights themselves are not adjusted by backpropagation. Instead, they behave as learning rate coefficients, which are updated after each epoch of training. 

\paragraph{Adaptive Resampling} In~\cite{wight2020solving}, a strategy to adaptively resample the residual collocation points based on the magnitude of the residual is proposed. While this approach improves the approximation, the training process must be interrupted and the MSE evaluated on the residual points to deterministically resample the ones with the highest error. After each resampling step, the number of residual points grows, increasing computational complexity. In \cite{tang2022adaptive}, resampling of the collocation points for solving the steady-state Fokker-Planck PDE is performed using an approximate density function, while in \cite{feng2021solving}, this work is extended to time-evolution problems by means of an adaptive density approximation method based on normalizing flows.



\paragraph{Neural Tangent Kernel (NTK) Weighting} Recently, \cite{wang2020and} derived the NTK kernel matrix for PINNs, and used a heuristic argument to set the weights adaptively based on the evolution of the eigenvalues of the NTK matrix during training.

\paragraph{Mimimax Weighting} In \cite{liu2021dual}, a methodology was proposed to update the weights during training using gradient descent for the network weights, and gradient ascent for the loss weights, seeking to find a saddle point in weight space. Loss components that do not decrease are assigned larger weights.

More generally, the need to use weighting to correct imbalance in multi-part loss functions has been recognized in the general deep learning literature~\cite{heydari2019softadapt,miranda2015multi,xu2018autoloss}. Note that the multi-part loss (\ref{eq:loss_weighted}) corresponds to a {\em linear scalarization} of this multiple-objective problem \cite{emmerich2018tutorial}.

All the previous methods employ a linearly-scalarized function such as (\ref{eq:loss_weighted}), the only difference among them being the way the weights are updated. The self-adaptive weighting method proposed in this paper is fundamentally different in that the weights apply to individual training points in the different loss components, rather than the entire loss component. The previous methods can be seen as a special case of this, when all self-adaptive weights for a particular loss component are updated in tandem. Among the previous methods, the independently-developed Minimax weighting scheme \cite{liu2021dual} is the closest to SA-PINNs, as it also  updates its weights via gradient ascent; however, these weights still apply to the whole loss components. This paper presents  empirical and theoretical evidence that having the flexibility of weighting each training point in the various loss terms brings additional flexibility that can lead to better performance.

\section{Self-Adaptive Physics-Informed Neural Networks}\label{sec:methods}

While previously proposed weighting methods produce improvements in stability and accuracy over the baseline PINN, they are either nonadaptive or introduce inflexible adaptation. Here we propose a simple procedure that applies fully-trainable weights to produce a multiplicative soft attention mask, in a manner that is reminiscent of attention mechanisms used in computer vision ~\cite{wang2017residual,pang2019mask}. Instead of hard-coding weights at particular regions of the solution, the proposed method is in agreement with the neural network philosophy of self-adaptation, where the weights in the loss function are updated by gradient descent side-by-side with the network weights.

\def\bl{\boldsymbol{\lambda}}
\def\l{\lambda}

Using the PDE in (1)-(3) as reference, the proposed self-adaptive PINN utilizes the following loss function 
\begin{equation}
    \mathcal{L}(\bw, \bl_r, \bl_b, \bl_{0}) = 
    \mathcal{L}_s(\bw) + \mathcal{L}_r(\bw,\bl_r) +  \mathcal{L}_{b}(\bw,\bl_b) + \mathcal{L}_{0}(\bw,\bl_0)\,, \label{eq:newloss}
\end{equation}
where $\bl_r = (\l^1_r,\ldots,\l_r^{N_r})$,  $\bl_b = (\l^1_b,\ldots,\l_b^{N_b})$, and $\mathbf{\bl}_{0} = (\l^1_{0},\ldots,\l^{N_0}_{0})$ are trainable, nonnegative {\em self-adaptation weights} for the initial, boundary, and residue points, respectively, and 
\begin{align}
    \mathcal{L}_r(\bw,\bl_r) &\,=\, \frac{1}{2}\sum^{N_r}_{i=1}m(\l^i_r)\, |\mathcal{N}_{\bx,t}[u(\bx^i_r,t^i_r;\bw)]-f(\bx^i_r,t^i_r)|^2 \label{Lr}\\
    \mathcal{L}_{b}(\bw,\bl_{b}) &\,=\, \frac{1}{2}\sum^{N_b}_{i=1}m(\l_{b}^i)\,
    |\mathcal{B}_{\bx,t}[u(\bx^i_r,t^i_r;\bw)]-g(\bx^i_b,t^i_b)|^2 \ \label{Lbt}\\  
    \mathcal{L}_{0}(\bw,\bl_{0}) &\,=\, \frac{1}{2}\sum^{N_0}_{i=1}m(\l_{0}^i)\,|u(\bx^i_0, 0;\bw)- h(\bx^i_0)|^2 \label{L0}.
\end{align}
where the {\em self-adaptation mask function} $m(\lambda)$ defined on $[0,\infty)$ is a nonnegative, differentiable on $(0,+\infty)$, strictly increasing function of $\lambda$. 
A key feature of self-adaptive PINNs is that the loss $\mathcal{L}(\bw, \bl_r, \bl_b, \bl_{0})$ is minimized  with respect to the network weights $\bw$, as usual, but is {\em maximized} with respect to the self-adaptation weights $\bl_r, \bl_b, \bl_0$. 
The corresponding gradient descent/ascent steps are:
\begin{align}
    \bw^{k+1} &\,=\, \bw^k \,-\, \eta_k\,\nabla_{\bw}\mathcal{L}(\bw^k, \bl^k_r, \bl^k_b, \bl^k_{0}) \label{eq-gdw}\\
    \bl_r^{k+1} &\,=\, \bl_r^k \,+\, \rho_r^k\nabla_{\bl_r}\mathcal{L}(\bw^k, \bl^k_r, \bl^k_b, \bl^k_{0}) \\
    \bl_b^{k+1} &\,=\, \bl_b^k \,+\, \rho^k_b\,\nabla_{\bl_b}\mathcal{L}(\bw^k, \bl^k_r, \bl^k_b, \bl^k_{0}) \\
    \bl_0^{k+1} &\,=\, \bl_0^k \,+\, \rho^k_0\,\nabla_{\bl_0}\mathcal{L}(\bw^k, \bl^k_r, \bl^k_b, \bl^k_{0})\,.
\end{align}
where $\eta^k>0$ is the learning rate for the neural network weights at step $k$, $\rho_p^k>0$ is a separate learning rate for the self-adaption weights, for $p=r,b,0$, and 

\begin{align}
    \nabla_{\bl_r}\mathcal{L}(\bw^k, \bl^k_r, \bl^k_b, \bl^k_{0}) &\,=\, \frac{1}{2}\begin{bmatrix}
        m^\prime(\l^{k,1}_r)\left|\mathcal{N}_{\bx,t}[u(\bx^i_r,t^i_r;\bw^k)]-f(\bx^1_r,t^1_r)\right|^2 \\ \cdots \\
        m^\prime(\l^{k,N_r}_r) \left|\mathcal{N}_{\bx,t}[u(\bx^i_r,t^i_r;\bw^k)]-f(\bx^{N_r}_r,t^{N_r}_r)\right|^2
        \end{bmatrix},\\  
    \nabla_{\bl_b}\mathcal{L}(\bw^k, \bl^k_r, \bl^k_b, \bl^k_{0}) &\,=\, \frac{1}{2}\begin{bmatrix}
        m^\prime(\l^{k,1}_b)\left|\mathcal{B}_{\bx,t}[u(\bx^i_b,t^i_b;\bw^k)]-g(\bx^1_b,t^1_b)\right|^2 \\ \cdots \\
        m^\prime(\l^{k,N_b}_b) \left|\mathcal{B}_{\bx,t}[u(\bx^i_b,t^i_b;\bw^k)]-g(\bx^{N_b}_b,t^{N_b}_b)\right|^2
        \end{bmatrix},\\
    \nabla_{\bl_0}\mathcal{L}(\bw^k, \bl^k_r, \bl^k_b, \bl^k_{0}) &\,=\, \frac{1}{2}\begin{bmatrix}
        m^\prime(\l^{k,1}_0)\left|u(\bx^1_0,0;\bw^k)-h(\bx^1_0,t^1_0)\right|^2 \\ \cdots \\
        m^\prime(\l^{k,N_0}_0) \left|u(\bx^i_0,0;\bw^k)]-h(\bx^{N_0}_0)\right|^2
        \end{bmatrix}.
\end{align}
Hence, since $m^\prime(\lambda) > 0$ (the mask function is strictly increasing, by assumption), then 
$\nabla_{\bl_r}\mathcal{L},\nabla_{\bl_b}\mathcal{L},\nabla_{\bl_0}\mathcal{L} \geq 0$, and any gradient component is zero if and only if the corresponding unmasked loss is zero. This shows that the sequences of weights $\{\bl^k_{r}; k=1,2,\ldots\}$, $\{\bl^k_{b}; k=1,2,\ldots\}$, $\{\bl^k_{0}; k=1,2,\ldots\}$ (and the associated mask values) are monotonically increasing, provided that the corresponding unmasked losses are nonzero. Furthermore, the magnitude of the gradients $\nabla_{\bl_r}\mathcal{L},\nabla_{\bl_b}\mathcal{L},\nabla_{\bl_0}\mathcal{L}$,
and therefore of the updates, are larger if the corresponding unmasked losses are larger. In addition, the magnitude of updates can be controlled by specifying a schedule for the learning rates $\rho^k_p$, for $p=r,b,0$, adding extra flexibility. This progressively penalizes the network more for not fitting the residual, boundary, and initial points closely.

We remark that any of the weights can be set to fixed, non-trainable values, if desired. For example, by setting $\l_b^k \equiv 1$, only the weights of the initial and residue points would be trained. The sensor data loss is not masked in this formulation, since if these data consist of noisy observations, weighting them requires extra care to avoid overfitting (though this remains an open research problem).
  
Notice that the self-adaptive weights need to be initialized at the beginning of training. They could be initialized to 1 (no weighting) or to a different value, depending on the problem. They could also be initialized randomly over an interval, similarly as to how neural network weights are often initialized. Here, prior knowledge plays an important role; e.g., if it is known that the initial conditions in a problem are hard to fit, then the initial condition weights could be initialized to a larger value than the other weights (alternatively, it could be initialized at the same value as the other weights, but employ a larger learning rate).

\begin{figure}[t!]
    \centering
    \includegraphics[width=.4\linewidth]{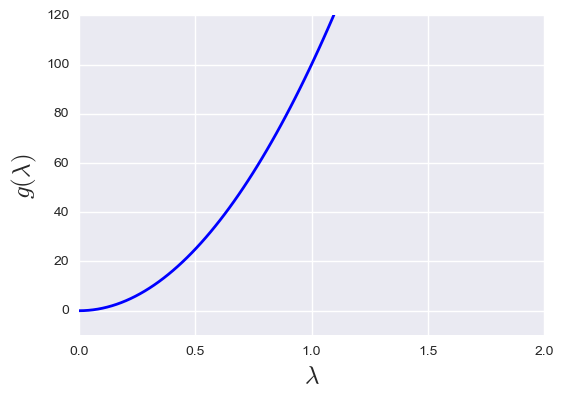}
    \includegraphics[width=.4\linewidth]{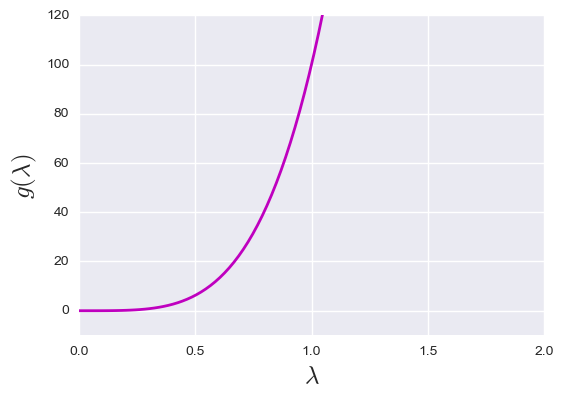}
    \includegraphics[width=.4\linewidth]{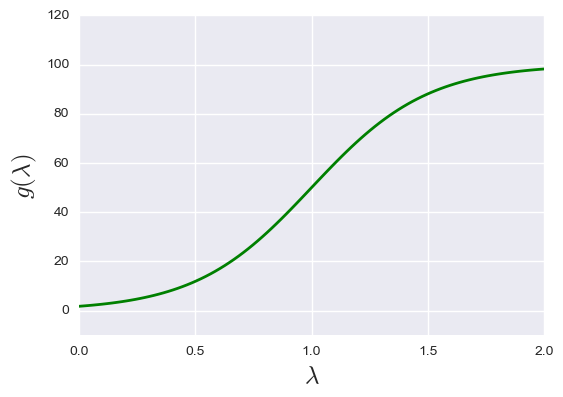}
    \includegraphics[width=.4\linewidth]{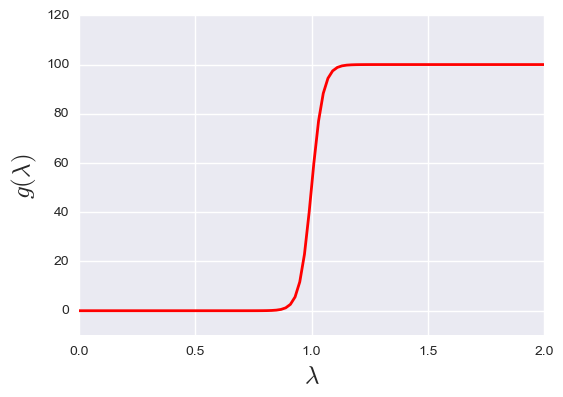}
    \caption{Mask function examples. From the upper left to the bottom right: polynomial mask, $q=2$; polynomial mask, $q=4$; smooth logistic mask; sharp logistic mask.
}
    \label{fig:masks}
\end{figure}

The shape of the function $g$ affects mask sharpness and training of the PINN. Examples include polynomial masks $m(\lambda) = c\lambda^q$, for $c,q>0$, and sigmoidal masks. See Figure~\ref{fig:masks} for a few examples. In practice, the polynomial mask functions have to be kept below a suitable (large) value, to avoid numerical overflow. The sigmoidal masks do not have this issue, and can be used to produce sharp masks. For example, in the bottom right example in Figure~\ref{fig:masks}, the mask is essentially binary; it starts small for small starting values of the self-adaptive weight $\lambda$, and after these exceed a certain threshold, the mask value will quickly take on the upper saturation value. Similarly to neural network nonlinearities, sigmoid mask functions can suffer from vanishing gradients during training. This is particularly a problem at the lower starting value. Therefore, excessively sharp sigmoidal mask functions should be avoided.

The gradient ascent/descent step can be implemented easily using off-the-self neural network software, by simply flipping the sign of $\nabla_{\bl_r}\mathcal{L}$, $\nabla_{\bl_b}\mathcal{L}$, and $\nabla_{\bl_0}\mathcal{L}$.
In our implementation of SA-PINNs, we use Tensorflow 2.3 with a fixed number of iterations of Adam~\cite{kingma2014adam}. In some case, these are followed by another fixed number of iterations of the L-BFGS quasi-newton method~\cite{liu1989limited}.  This is consistent with the baseline PINN formulation in~\cite{raissi2019physics}, as well as follow-up literature~\cite{wight2020solving}.
However, the adaptive weights are only updated in the Adam training steps, and are held constant during L-BFGS training, if any. A full implementation of the methodology described here has been made publicly available by the authors\footnote{https://github.com/levimcclenny/SA-PINNs} and it is included in the open-source software \textit{TensorDiffEq}~\cite{mcclenny2021tensordiffeq}.
Finally, we remark that there are some similarities between SA-PINN training and penalty methods in optimization, which introduce a sequence of increasing penalty costs~\cite{luenberger2008linear}.

\section{Numerical Examples}
\label{sec:results}

In this section we present numerical experiments demonstrating the SA-PINN performance on various benchmarks. The main figure of merit used is the L2-error:
\begin{equation}
    L_2 \ \text{error} \,=\, \frac{\sqrt{\sum_{i=1}^{N_U} |u(x_i,t_i) - U(x_i,t_i)|^2}}{\sqrt{\sum_{i=1}^{N_U} |U(x_i,t_i) |^2}}\,.
\end{equation}
where $u(x,t)$ is the trained approximation, and $U(x,t)$ is a high-fidelity solution over a fine mesh $\{x_i,t_i\}$ containing $N_U$ points. In all cases below, we repeat the training process over 10 random restarts and report the average L2 error and its standard deviation.


\subsection{Viscous Burgers Equation}

The viscous Burgers PDE considered here is 
\begin{align}
&u_t + uu_x - (0.01/\pi) u_{xx} = 0\,,  \ \  x \in [-1,1], \ t \in [0,1]\,,\\
&u(0,x) = -\sin(\pi x) \,,\\
&u(t, -1) = u(t, 1) = 0\,.
\end{align}

All results for the viscous Burgers PDE were generated from a fully-connected network with input layer size 2 corresponding to the $x$ and $t$ inputs, 8 hidden layers of 20 neurons each, and an output layer of size 1 corresponding to the output of the approximation $u(x,t)$. This directly mimics the setup of the viscous Burgers PDE result presented in~\cite{raissi2019physics}. All training is done for 10k iterations of Adam, followed by 10k iterations of L-BFGS to fine tune the network weights, consistent with related work. Additionally, the number of points selected for the trials shown are $N_0$ = 100, $N_b$ = 200, and $N_r$ = 10000. Training with this architecture took 96ms/iteration on a single Nvidia V100 GPU. We initialize the self adaptive weights on the IC and the residual points to be $U(0,1)$ and the learning rates for all self-adaptive weights were set to $5e\!-\!3$.

\begin{figure}[hbt!]%
    \centering
    \subfloat{\includegraphics[width=.49\linewidth]{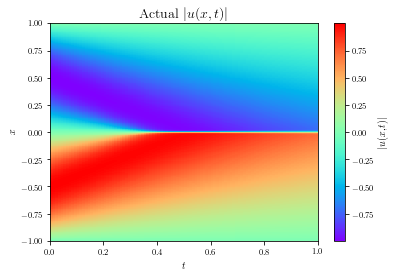} }
    \subfloat{{\includegraphics[width=.49\linewidth]{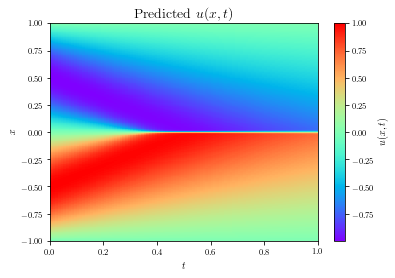} }}%
    \caption{High-fidelity (\textit{left}) vs. predicted (\textit{right}) solutions for the viscous Burgers PDE.}%
    \label{fig:burg_pred}%
\end{figure}

\begin{figure}[hbt!]
\centering
\includegraphics[width=.7\linewidth]{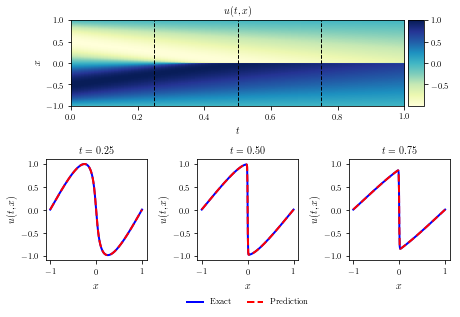}\\
\includegraphics[width=0.49\linewidth]{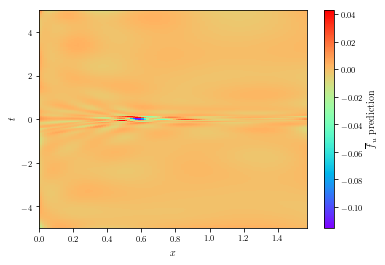}
\includegraphics[width=0.49\linewidth]{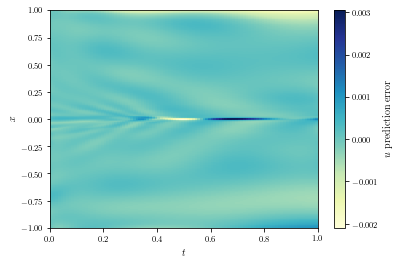}
\caption{\textit{Top:} predicted solution of the viscous Burgers PDE. \textit{Middle:} Cross-sections of the approximated vs. actual solutions for various x-domain snapshots. \textit{Bottom left:} Residual $r(x,t)$ across the spatial-temporal domain. \textit{Bottom right:} Absolute error between prediction and high-fidelity solution across the spatial-temporal domain.
}
\label{fig:burg_trained}
\end{figure}

We achieved an L2-error of $4.80e\!-\!4 \pm 1.01e-4$, which is smaller value than the $6.7e\!-\!4$ L2 error reported in~\cite{raissi2019physics}, while using only 20\% as many training iterations and an identical neural network  architecture. The high-fidelity and predicted solutions  are displayed in figure~\ref{fig:burg_pred}. 
Figure~\ref{fig:burg_trained} demonstrate the accuracy of the proposed approach, using a significantly shorter training horizon than the baseline PINN. 

\begin{figure}[hbt!]
\centering
\includegraphics[width=.6\linewidth]{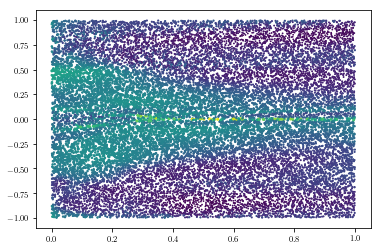}
\caption{Trained weights for residue points across the domain $\Omega$. Larger/brighter colored points correspond to larger weights. 
}
\label{fig:burg_weights}
\end{figure}

Figure~\ref{fig:burg_weights} shows that the sharp discontinuity at $x=0$ in the solution has correspondingly large weights, indicating that the model must pay extra attention to those particular points in its solution, resulting in an increase in approximation accuracy and training efficiency.

\subsection{Helmholtz Equation}

The Helmholtz PDE is typically used to describe the behavior of wave and diffusion processes, and can be employed to model evolution in a spatial domain or combined spatial-temporal domain. Here we study a particular Helmholtz PDE existing only in the spatial $(x,y)$ domain, described as:
\begin{align}
& u_{xx} + u_{yy} + k^2u - q(x,y) = 0 \label{eq:helm_sys} \\
& u(-1,y) = u(1,y) = u(x,-1) = u(x,1) = 0 \label{eq:helm_bound}
\end{align}
where $x \in [-1,1], y \in [-1,1]$ and
\begin{align}
q(x,y) = -\  &(a_1\pi)^2\sin(a_1\pi x)\sin(a_2\pi y) \ \nonumber\\
- \ &(a_2\pi)^2\sin(a_1\pi x)\sin(a_2\pi y) \ \nonumber\\
+ \ &k^2\sin(a_1\pi x)\sin(a_2\pi y) \label{eq:helm_forcing}
w\end{align}
is a forcing term that results in a closed-form analytical solution
\begin{equation}
    u(x,y) = \sin(a_1\pi x)\sin(a_2\pi y)\,.
\end{equation}
To allow a direct comparison to the Helmholtz PDE result reported in~\cite{wang2020understanding}, we take $a_1$ = 1 and $a_2$ = 4 and use the same neural network architecture with layer sizes [2, 50, 50, 50, 50, 1]. Our architecture is trained for 10k Adam and 10k L-BFGS iterations, again keeping the self-adaptive mask weights constant through the L-BFGS training iterations and only allowing those to train via Adam. We sample $N_b$ = 400 (100 points per boundary). Given the steady-state initialization and constant forcing term, there is no applicable initial condition and consequently no $N_0$. We create a mesh of size (1001,1001) corresponding to the $x \in [-1,1], y \in [-1,1]$ range, yielding 1,002,001 total mesh points, from which we select $N_r$=100k residue points. We initialize the self adaptive weights on the BC and the residual points to be $U(0,1)$ and the learning rates for all self-adaptive weights were set to 5e-3. 

\begin{figure}[hbt!]%
    \centering
    \subfloat{{\includegraphics[width=.49\linewidth]{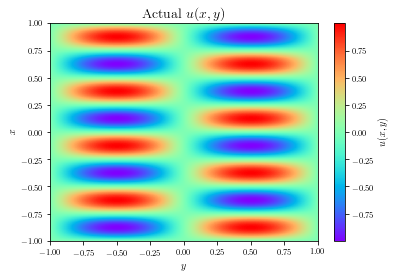} }}
    \subfloat{{\includegraphics[width=.49\linewidth]{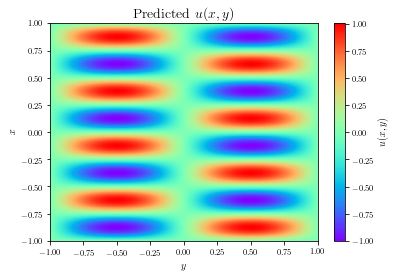} }}%
    \caption{Exact (\textit{left}) vs. predicted (\textit{right}) solutions for the Helmholtz PDE.}%
    \label{fig:helm_example}
\end{figure}

We can see in figure~\ref{fig:helm_example} that the SA-PINN prediction is very accurate and indistinguishable from the exact solution, with an L2 error of  $3.2e\!-\!3 \pm 2.2e\!-\!4$. With a larger neural network, the results reported in in~\cite{wang2020understanding} (row 5 of Table~2) are $1.4e\!-\!1$ for the baseline PINN, and between $2.54e\!-\!3$ and $2.74e\!-\!2$ for various learning-rate annealing weighted schemes proposed in that paper. We would add that the SA-PINN is trained for a smaller number of iterations (10k Adam and 10k L-BFGS) with respect to that in~\cite{wang2020understanding} (40k Adam).

\begin{figure}[hbt!]
\centering
\includegraphics[width=.7\linewidth]{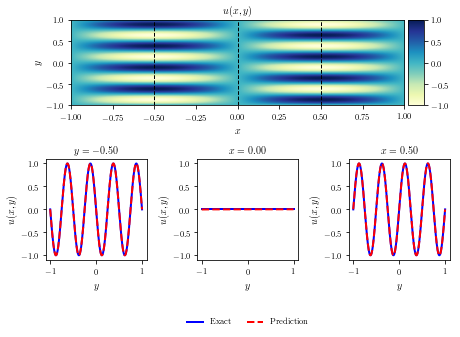}\\
\includegraphics[width=0.49\linewidth]{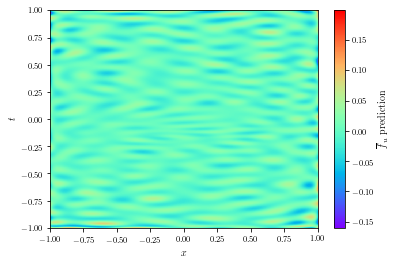}
\includegraphics[width=0.49\linewidth]{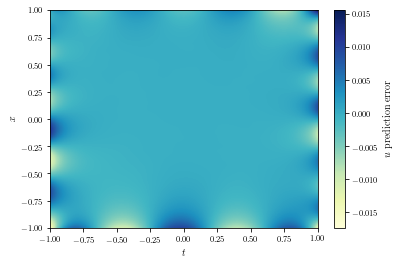}
\caption{\textit{Top} predicted solution of Helmholtz equation. \textit{Bottom} Cross-sections of the approximated vs. actual solutions for various x-domain snapshots
}
\label{fig:helm_trained}
\end{figure}

\begin{figure}[hbt!]
\centering
\includegraphics[width=.6\linewidth]{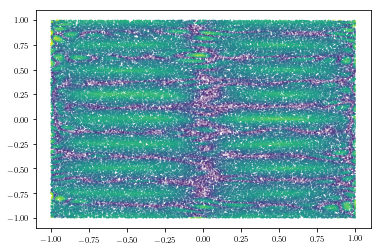}
\caption{Self-learned weights after training via Adam for the Helmholtz system. Brighter/larger points correspond to larger weights.
}
\label{fig:helm_weights}
\end{figure}

Figure~\ref{fig:helm_trained} shows individual cross-sections of the Helmholtz solution, demonstrating the SA-PINN's ability to accurately approximate the sinusoidal solution on the whole domain. Figure~\ref{fig:helm_weights} shows that the Self-Adaptive PINN largely ignores the flat areas in the solution, while focusing its attention on the nonflat areas. 

\subsection{Allen-Cahn Reaction-Diffusion Equation}\label{sec:AC}

In this section, we report experimental results obtained with the Allen-Cahn PDE, which contrast the performance of the proposed SA-PINN algorithm against the baseline PINN and two of the PINN algorithms mentioned in Section \ref{Sec:rw}, namely, the nonadaptive weighting and time-adaptive schemes (for the latter, Approach 1 in \cite{wight2020solving} was used).


The Allen-Cahn reaction-diffusion PDE is typically encountered in phase-field models, which can be used, for instance, to simulate the phase separation process in the microstructure evolution of metallic alloys~\cite{moelans2008introduction,shen2010numerical,kunselman2020semi}. The Allen-Cahn PDE considered here is specified as follows:
\begin{align}
   &u_t -0.0001u_{xx} + 5u^3 -5u = 0\,, \ \  x \in [-1,1], \ t \in [0,1]\,,\\
   & u(x, 0) = x^2cos(\pi x)\,, \\
    & u(t, -1) = u(t,1)\,, \label{BC1} \\
   & u_x(t,-1) = u_x(t,1)\,. \label{BC2}
\end{align}

The Allen-Cahn PDE is an interesting benchmark for PINNs for multiple reasons. It is a ``stiff'' PDE that challenges PINNs to approximate solutions with sharp space and time transitions, and is also introduces periodic boundary conditions (\ref{BC1},~\ref{BC2}). 
In order to deal with the latter, 
the boundary loss function $\mathcal{L}_{b}(\bw,\bl_{b})$ in (\ref{Lbt}) is replaced by
\begin{equation}
    \mathcal{L}_{b}(\bw,\bl_{b}) \,=\, \frac{1}{N_b}\sum^{N_b}_{i=1}g(\l_{b}^i)(|u(1, t^i_b) - u(-1, t^i_b)|^2 + |u_x(1, t^i_b) - u_x(-1, t^i_b)|^2)\,
\end{equation}

The neural network architecture is fully connected with layer sizes \linebreak$[2, 128, 128, 128, 128, 1]$. This architecture is identical to the one used in the Allen-Cahn PDE result reported in~\cite{wight2020solving}, in order to allow a direct comparison of performance. We set the number of residue, initial, and boundary points to $N_r = 20,000, N_0 = 100$ and  $N_b = 100$, respectively (due to the periodic boundary condition, there are in fact 200 boundary points). Here we hold the boundary weights $w^i_b$ at 1, while 
the initial weights $w^i_0$ and residue weights $w^i_r$ are trained. The initial and residue weights are initialized from a uniform distribution in the intervals $[0,100]$ and $[0,1]$, respectively. Training took 65ms/iteration on a single Nvidia V100 GPU. 

\begin{figure}[t!]
    \centering
    \includegraphics[width=1\linewidth]{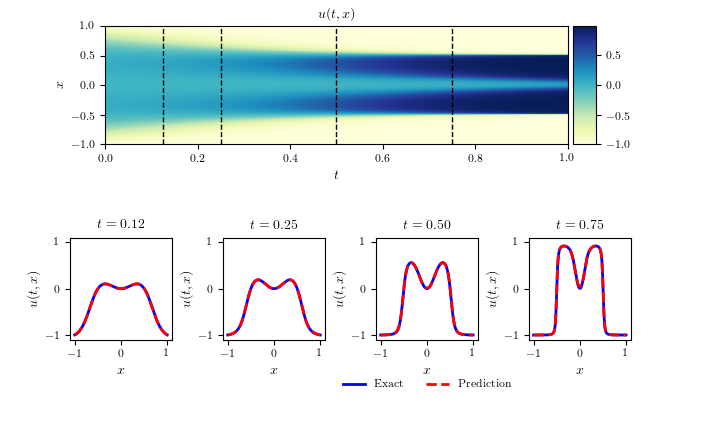}\\
    \includegraphics[width=0.49\linewidth]{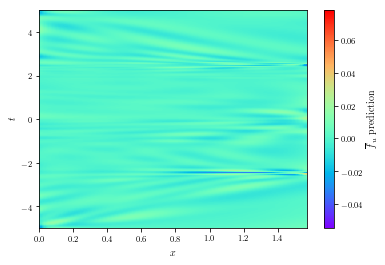}
    \includegraphics[width=0.49\linewidth]{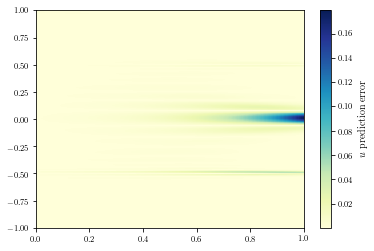}
    \caption{\textit{Top:} Plot of the approximation $u(x,t)$ via the SA-PINN. \textit{Middle:} Snapshots of the approximation $u(x,t)$ vs.\ the high-fidelity solution $U(x,t)$ at various time points through the temporal evolution. \textit{Bottom left:} Residual $r(x,t)$ across the spatial-temporal domain. As expected, it is close to 0 for the whole domain $\Omega$. \textit{Bottom right:} Absolute error between approximation and high-fidelity solution across the spatial-temporal domain.
}
    \label{fig:trained}
\end{figure}

Numerical results obtained with the SA-PINN are displayed in figure~\ref{fig:trained}. The average L2-error across 10 runs with random restarts was $2.1e\!-\!2 \pm  1.21e\!-\!2$, while the L2-error on 10 runs obtained by the time-adaptive approach in \cite{wight2020solving} was $8.0e\!-\!2 \pm 0.56e\!-\!2$. Neither the baseline PINN nor the nonadaptive weighted scheme, with initial condition weight $C=100$, were able to solve this PDE satisfactorily, with L2 errors $96.15e\!-\!2 \pm 6.45e\!-\!2$ and $49.61e\!-\!2 \pm 2.50e\!-\!2$, respectively (these numbers matched almost exactly those reported in \cite{wight2020solving}).

\begin{figure}[hbt!]
    \centering
    \includegraphics[width=.6\linewidth]{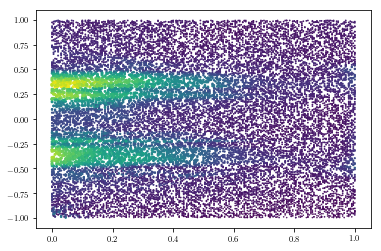}
    \caption{Learned self-adaptive weights across the spatio-temporal domain. Brighter colors and larger points indicate larger weights.}
    \label{fig:weights}
\end{figure}

Figure~\ref{fig:weights} is unique to the proposed SA-PINN algorithm. It displays the trained self-adaptive weights for the residue points across the spatio-temporal domain. These are the weights of the multiplicative soft attention mask self-imposed by the PINN. This plot stays remarkably constant across different runs with random restarts, which is an indication that it is a property of the particular PDE being solved. We can observe that in this case, more attention is needed early in the solution, but not uniformly across the space variable. In \cite{wight2020solving}, this observation was justified by the fact that the Allen-Cahn PDEs describes a time-irreversible diffusion-reaction processes, where the solution has to be approximated well early. However, here this fact is ``discovered'' by the SA-PINN itself.

\begin{figure}[hbt!]
    \centering
    \includegraphics[width=\linewidth]{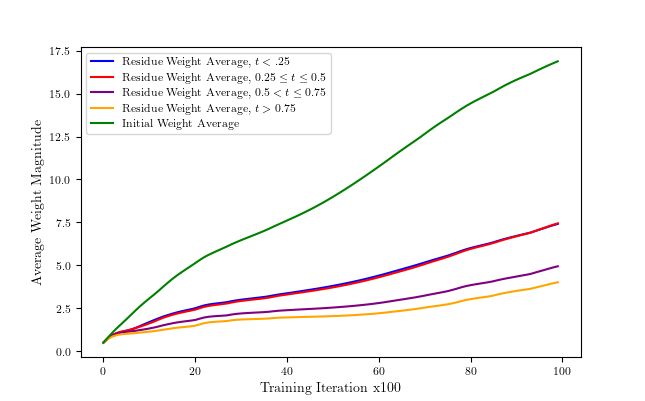}
    \caption{Average learned residue weights across various partitions of the solution domain. Note that earlier times require heavier weighting, with the highest average weights being the initial condition weights. This is consistent with the rationale that earlier solutions must be correctly learned for time-diffusive processes. }
    \label{fig:weights_avg}
\end{figure}

In order to study the behavior of the SA-PINN more closely, 
we plot in Figure~\ref{fig:weights_avg} the average value of the residue weights from various partitions of the solution domain. While all the weights are increasing, as must be the case since the mask function is required to be monotone, the rate of increase is of importance. Notice that the initial condition weights grow much faster than the residue weights, as expected, since the initial condition tends to be neglected by the PINN, otherwise. As for the residue weights, we see that, for small values of $t$, they increase faster than for large values of $t$. This shows that the SA-PINN has learned that the early part of the evolution is the most critical part of the solution. (This agrees with what was seen in the map of Figure~\ref{fig:weights}.) In contrast with traditional time-marching approaches, where earlier time steps are solved prior to later ones, the SA-PINN solves the PDE over the entire space-time domain at once; however, the self-adaptive weights allow it to concentrate on the early part of the evolution.

\begin{figure}[t!]
    \centering
    \includegraphics[width=.58\linewidth]{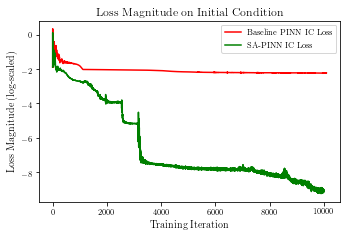}
    \includegraphics[width=.58\linewidth]{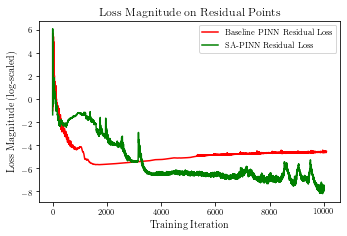}
    \includegraphics[width=.58\linewidth]{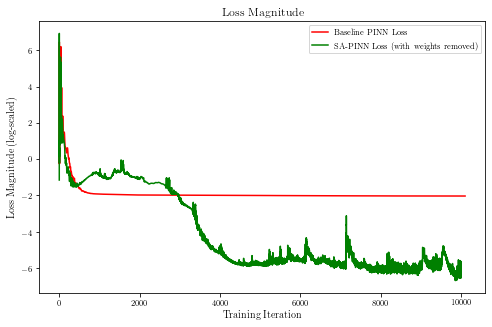}    
    \caption{Average values for the initial condition loss, residue loss, and total loss, over 10k Adam training iterations. For the SA-PINN, the weights were removed from the loss value to provide a direct comparison to the baseline.}
    \label{fig:IC_loss}
\end{figure}

Finally, Figure~\ref{fig:IC_loss} displays the training loss for the baseline PINN and SA-PINN as a function of training iteration. For the SA-PINN, the weights were removed from the loss value to provide a direct comparison to the baseline. These plots are generated from 10 random restarts of the SA and baseline PINN training cycles over 10k Adam training iterations with consistent learning rates. We can see that the SA-PINN achieves significantly lower initial condition loss than the baseline PINN for the initial. Indeed, this is the major issue faced by the baseline PINN in the AC problem. As for the residual loss, we see that the baseline PINN decreases it fast (at the expense of the initial condition loss), but that eventually the SA PINN is able to achieve a loss two orders of magnitude smaller. The oscillatory behavior of the residue loss in the SA-PINN reveals the dynamics of the competing self-adaptive weighted initial condition and residue loss terms.

\subsection{2D Burgers Equation}

\begin{figure}[hbt!]
    \centering
    \begin{tabular}{cc}
    \includegraphics[width=.43\linewidth]{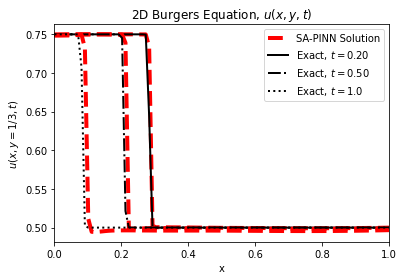} & \hspace{-1em}
    \includegraphics[width=.43\linewidth]{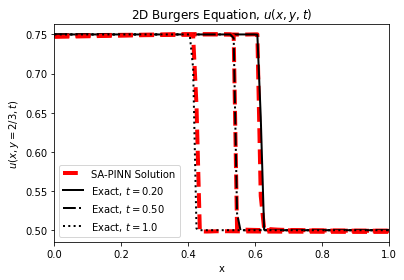}\\[-1ex]
    \includegraphics[width=.43\linewidth]{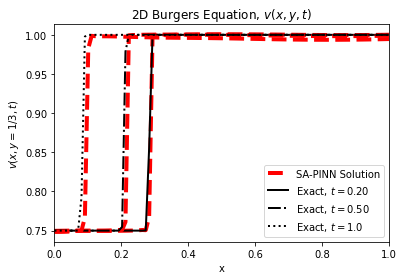} & \hspace{-1em}
    \includegraphics[width=.43\linewidth]{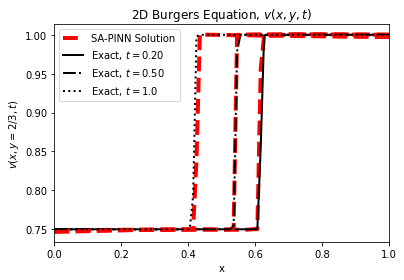} \\[-1ex]
\end{tabular}
    \caption{SA-PINN solution of 2D Burgers equation example. The L2 error for both $u$ and $v$ at $t=1/3$ and $t=2/3$ is approximately $3e\!-\!3$.}
    \label{fig:2d-quad}
\end{figure}

Here we demonstrate the efficacy of SA-PINN in the solution of a three-dimensional problem (two spatial dimensions plus time), namely, a 2D viscous Burgers nonlinear PDE system with velocity fields $u(x,y,t)$ and $v(x,y,t)$, which satisfy:
  \begin{align}
  &u_t + uu_x + vu_y= \nu (u_{xx}+u_{yy})\,,\\
  &v_t + uv_x + vv_y= \nu (v_{xx}+v_{yy})\,, 
  \end{align}
in the domain $(x,y,t) \in (0,1)^3$, where the $\nu$ is the kinematic viscosity (in this example, $\nu = 0.002$). Any velocity fields such that $u(x,y,t)+v(x,y,t)$ is constant provide a solution of this PDE system. Following~\cite{lu2021deepxde}, we take:
\begin{align}
    u(x,y,t) & = \frac{3}{4} - \frac{1}{4}\left[1+\exp\left(\frac{-4x + 4y -t}{32\nu}\right)\right]^{-1}, \\
   v(x,y,t) & = \frac{3}{4} + \frac{1}{4}\left[1+\exp\left(\frac{-4x + 4y -t}{32\nu}\right)\right]^{-1},
   \label{eq-burgers2}
\end{align}
with matching initial and Dirichlet boundary conditions. We apply the SA-PINN without enforcing the boundary conditions, i.e., only the initial conditions are enforced.
Despite this, the SA-PINN was able to capture the attenuated shock effectively as shown in Figure~\ref{fig:2d-quad}. These results were obtained with 35k Adam iterations at a learning rate of $1e\!-\!5$, followed by 20k L-BFGS iterations to fine-tune the solution. Training of the baseline PINN for the same amount amount of iterations was not successful, as the L-BFGS optimizer failed consistently to converge. Notably, SA-PINN appears to stabilize the training process, allowing the L-BFGS optimizer to converge.

\subsection{Self-Adaptive Weight Training Hyperparameters}

We comment here on the choice of hyperparameter settings made in the experiments reported in the previous sections. In all experiments, 
a constant learning rate of 5e-3 for the gradient ascent of the self-adaptive learning rates, and Adam optimization was used. Better results could potentially be obtained by using learning rate scheduling.

Empirically, we observed that effective training strategies for self-adaptive PINNs tend to require smaller values of learning rate for the neural network weights (i.e., 1e-5), and larger values  for the self-adaptive weights (i.e., 1e-3 to 1e-1). 

As specified in Section~\ref{sec:AC}, the results shown in Figure~\ref{fig:trained} employ random initialization of the initial condition and residue self-adaptive weights in the intervals $[0,100]$ and $[0,1]$, respectively, and the learning rates are held constant and equal. This choice is dictated by the prior knowledge that heavier weighting of the initial condition is needed in the AC problem~\cite{wight2020solving}. On the other hand, in the results displayed in figure~\ref{fig:weights_avg}, all weights were initialized randomly in the interval $[0,1]$. In this case, it is observed that the initial conditions increase faster on their own, even though all learning rates are held constant and equal.

\section{Self-Adaptive PINNs with Stochastic Gradient Descent}

Stochastic gradient descent (SGD) \cite{robbins1951stochastic} uses randomly sampled subsets of the training data to compute approximations to the gradient for training neural networks by gradient descent \cite{ruder2016overview}. It has been claimed that the empirical superior performance of stochastic gradient descent over large-batch training is due to a tendency of the latter to converge to ``sharp'' minima in the loss surface, which have poor performance, while SGD with small batches converge to better ``flat'' minima~\cite{keskar2016large}. 

The issue has not been well studied in the context of PINNs at the time of writing, though there is some empirical evidence that SGD can indeed improve the $L_2$ performance of PINNs with some PDEs. It should be pointed out that PINNs are well-suited to SGD since a new set of residue, initial and boundary points can be sampled each time rather than subsampling a given set of training data points as in conventional machine learning. 

The baseline SA-PINN algorithm described previously cannot take advantage of small-batch SGD since the self-adaptive weights are attached to specific training points. In this section, we examine an extension of SA-PINN that allows the use of SGD. The basic idea is to use a spatial-temporal predictor of the value of self-adaptive weights for the newly sampled points. Here we use standard Gaussian process regression due to its predictive power. (However, simpler regression approaches could be equally used, in cases where GP regression is unwieldy, e.g., due to large sample size.)

A problem where SGD has been found empirically to have a strong impact is the 1D wave equation:
\begin{align}
    & u_{tt}(x, t) - 4u_{xx}(x, t)  \,=\, 0\,, \ \ x \in [0,1]\,, \ t \in [0,1]\,, \\
    & u(0,t) = 0 \,, u(1,t) = 0, \ t \in [0,1]\,,\\
    & u_t(x,0) = 0 \,, \ x \in [0,1]\,,\\
    & u(x,0) = \sin(\pi x) + \frac{1}{2}\sin(4\pi x)\,, \ x \in [0,1]\,.
\end{align}
This problem was considered in \cite{wang2020and} to study their NTK weighting scheme. The problem has an analytical solution:
\begin{equation}
     u(x,t) \,=\, \sin(\pi x) \cos(2\pi t) + \frac{1}{2}  \sin(4\pi x) \cos(8\pi t)\,, \ \ x \in [0,1]\,, \ t \in [0,1]\,.
\end{equation}       

The baseline PINN struggles in this problem due to its stiffness. Here, we investigate the improvement provided by SGD, fixed weights, and self-adaptive weights. The architecture of the neural network consists of 5 layers of 500 neurons each with the tanh nonlinearity, and the number of residue, initial, and boundary points were set to 300, 100, and 100, respectively (these are the same hyperparameters used in \cite{wang2020and}). 
The small sample sizes are appropriate to study the impact of SGD. 

In all experiments, the learning rate for the neural network weights is kept fixed at $10^{-5}$ for a total of 80,000 iterations. The self-adaptive weights are all initialized to 1.0, with learning rate 0.01 for the residue points, 0.05 for the  initial condition on $u_t$, and 0.25 for all other initial and boundary conditions on $u$.  
In the fixed-weight experiment, the weights were kept constant at 1.0 for the residue points, 5.0 for the  initial condition on $u_t$, and 50.0 for all other initial and boundary conditions on $u$. These values make the final average values taken by the self-adaptive weights at the end of training approximately match the fixed weights. SGD is applied by resampling all training points every 100 iterations. The GPs were trained using fixed hyperparameters (no automatic tuning is performed). 

Results based on 10 independent random initializations of the neural network weights are displayed in Table~\ref{table:SGD}. We can observe that all methods fail in the absence of SGD. On the other hand, while SGD is not able to improve the performance of the baseline PINN, it produces a significant improvement to the fixed-weight PINN, and a large improvement to the SA-PINN. In fact, the SA-PINN achieves an average L2-error of 2.95\%, which is an order of magnitude better than the fixed-weight result. This L2-error is however larger than the one reported in \cite{wang2020and}. Optimizations to the SGD process, including adaptive tuning of the GP hyperparameters, will be part of future work, and are expected to improve performance.

\begin{figure}[hbt!]
    \centering
    \begin{tabular}{cc}
    \multicolumn{2}{c}{\includegraphics[width=.40\linewidth]{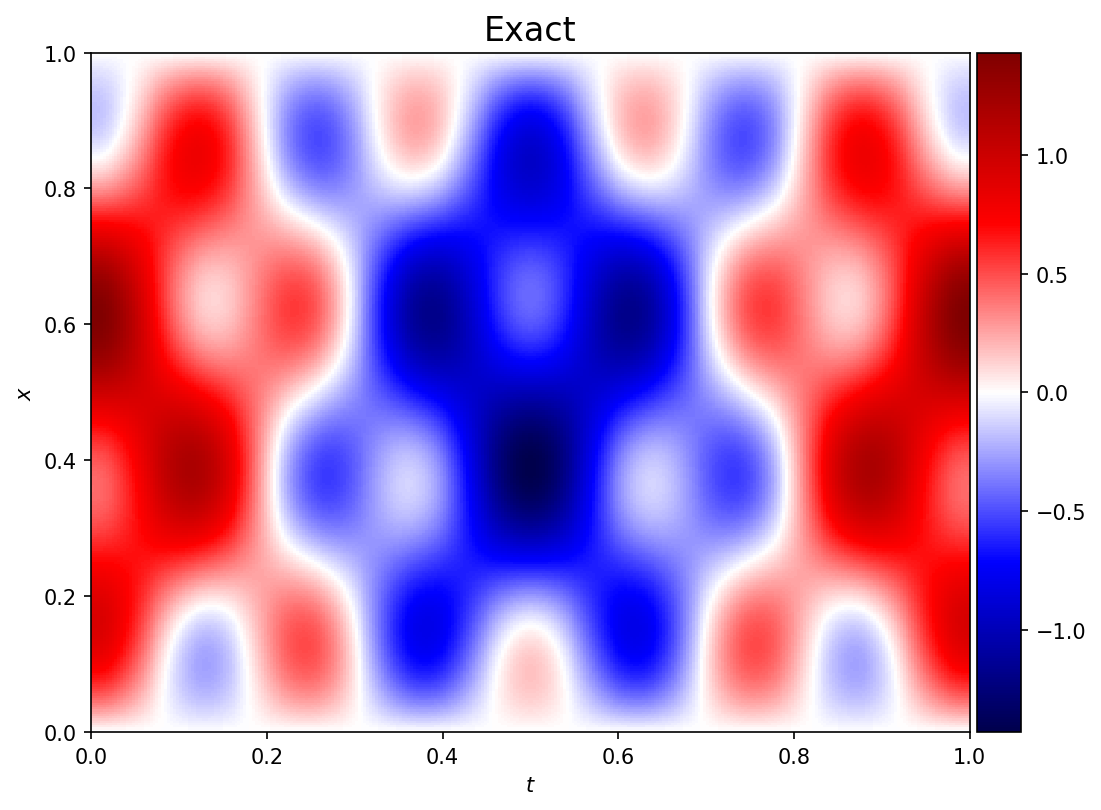}}\\[-1ex]
    \includegraphics[width=.40\linewidth]{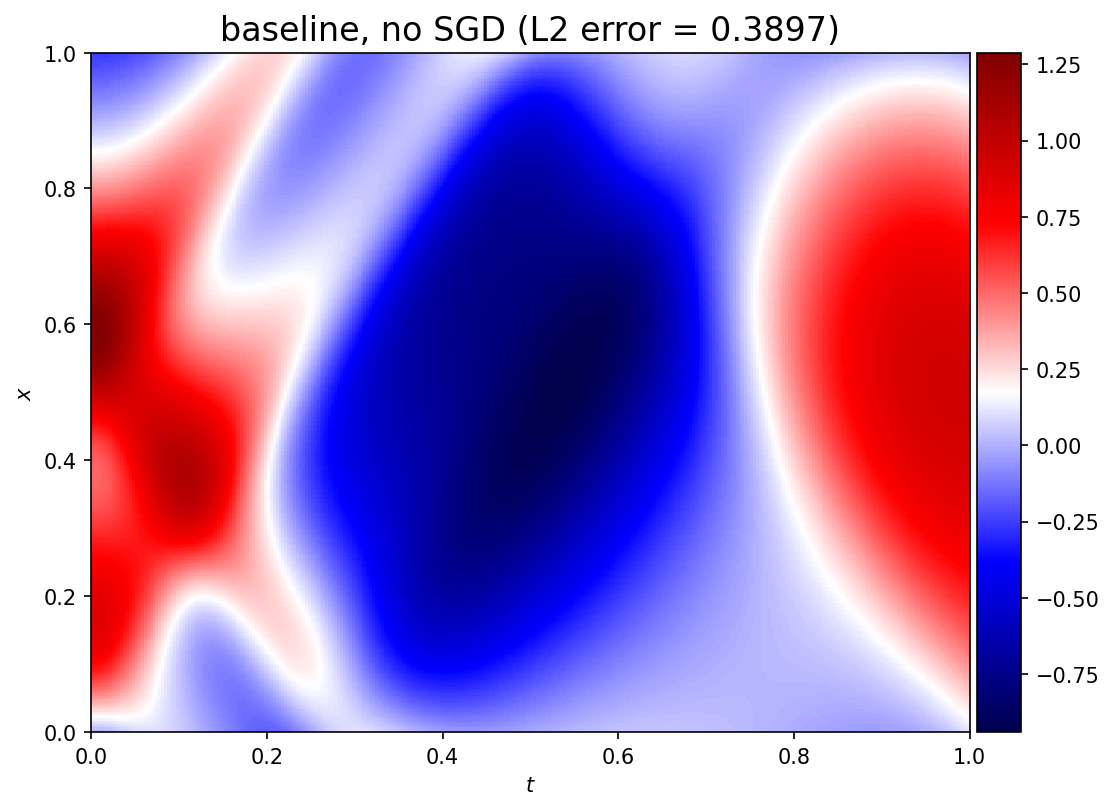} & \hspace{-1em}
    \includegraphics[width=.40\linewidth]{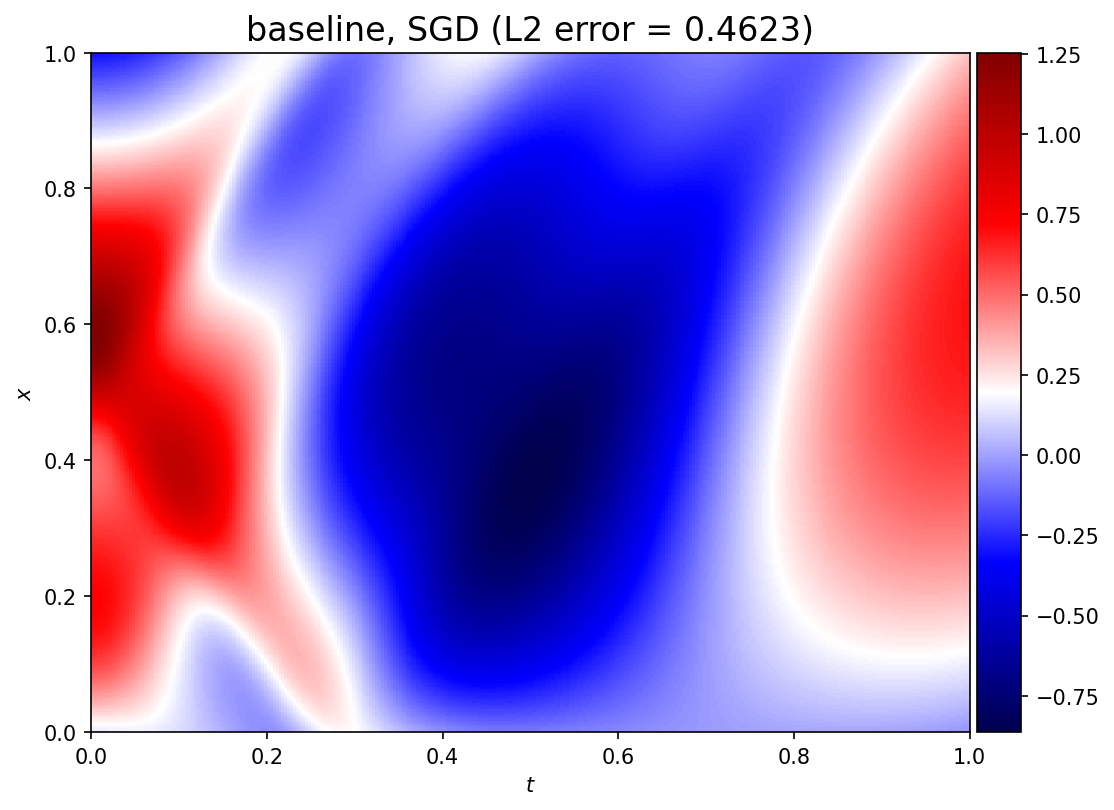}\\[-1ex]
    \includegraphics[width=.40\linewidth]{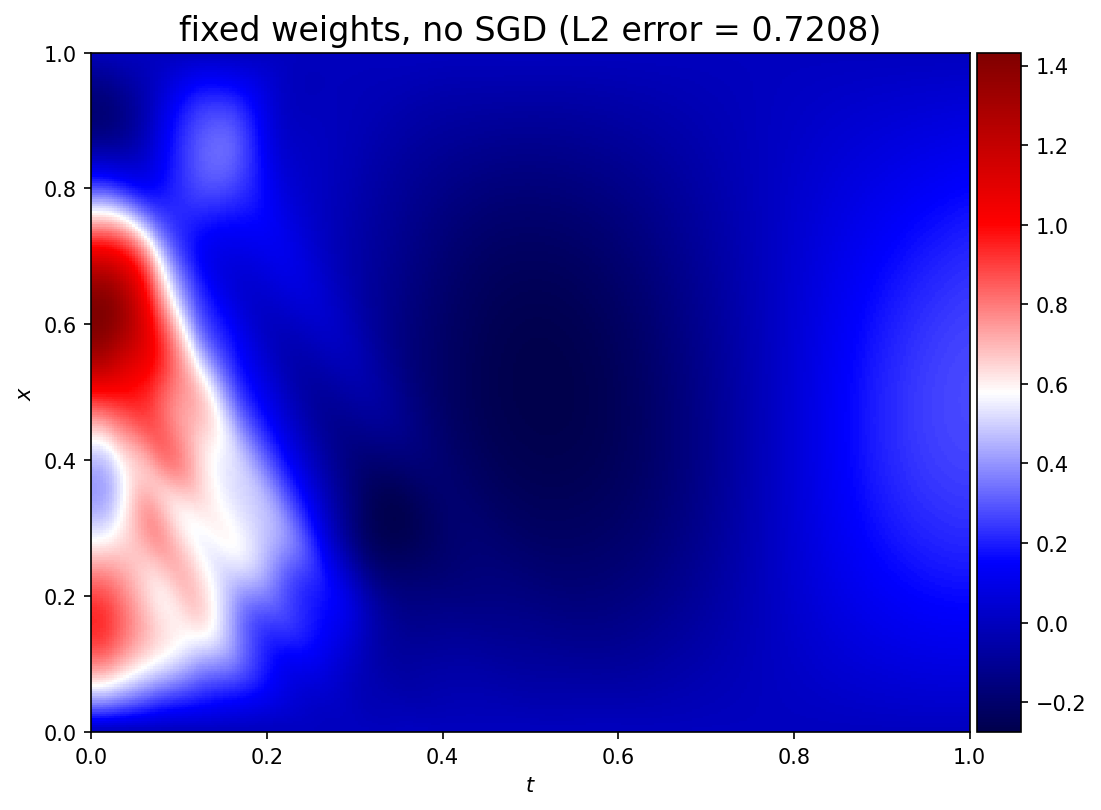} & \hspace{-1em}
    \includegraphics[width=.40\linewidth]{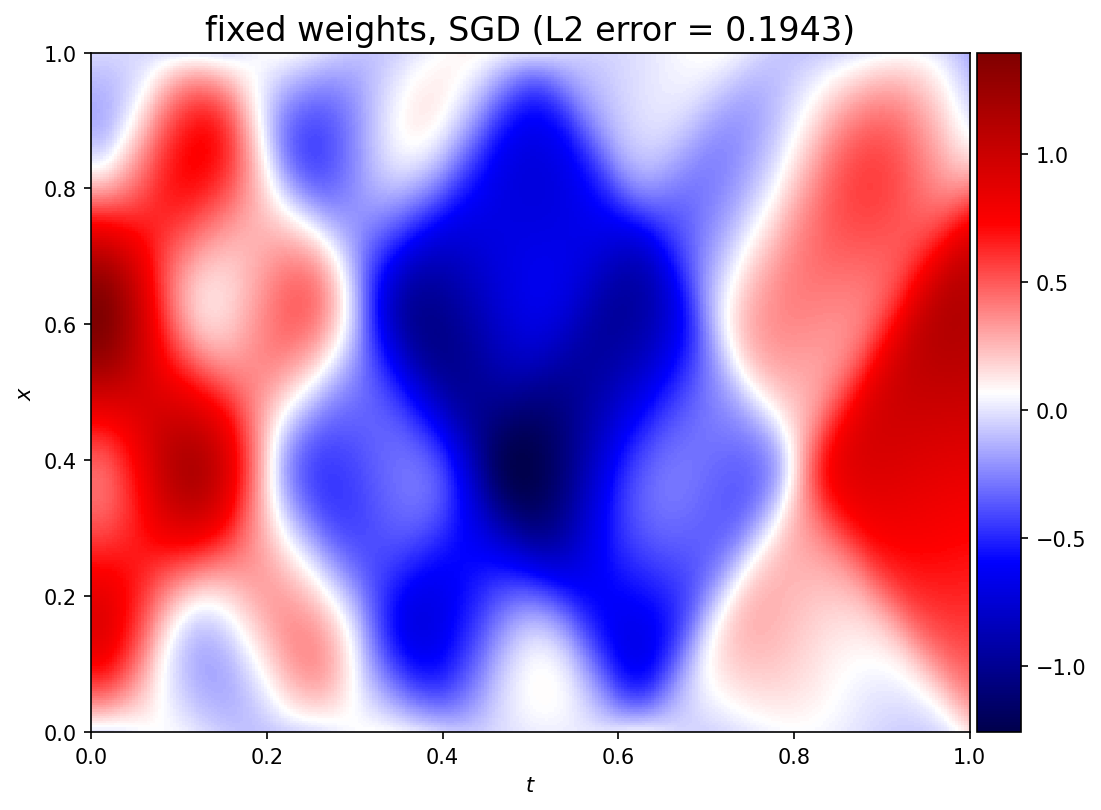} \\[-1ex]
    \includegraphics[width=.40\linewidth]{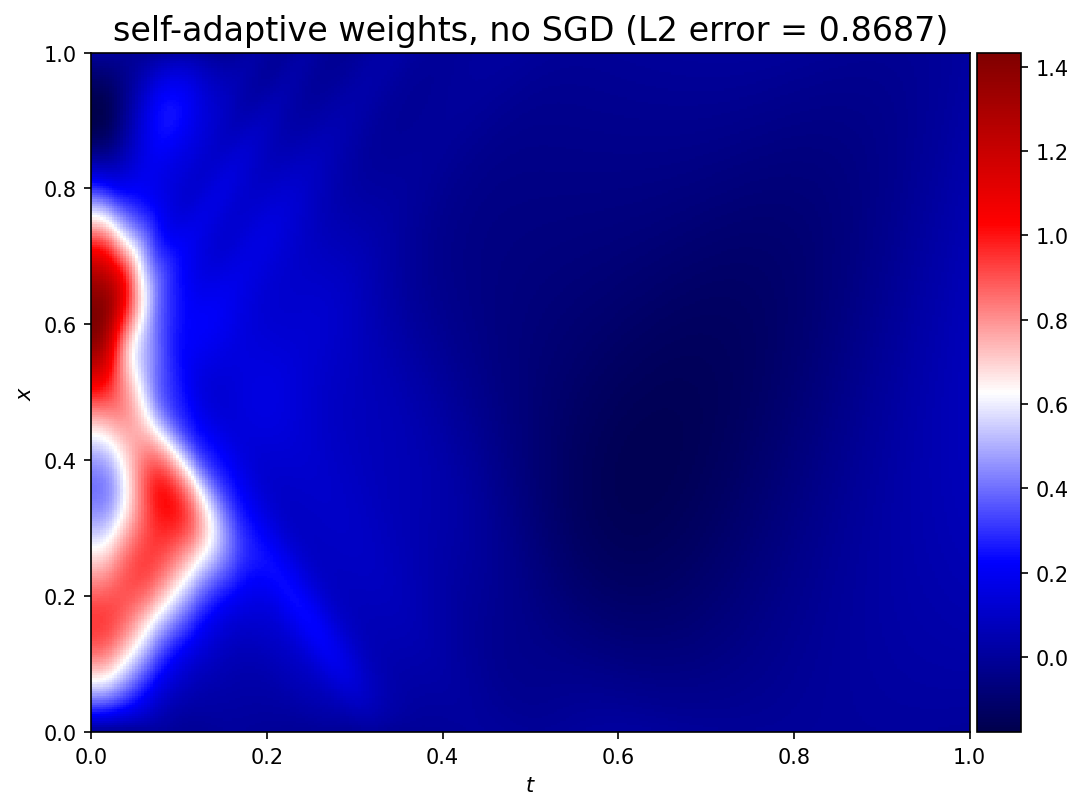} & \hspace{-1em}
    \includegraphics[width=.40\linewidth]{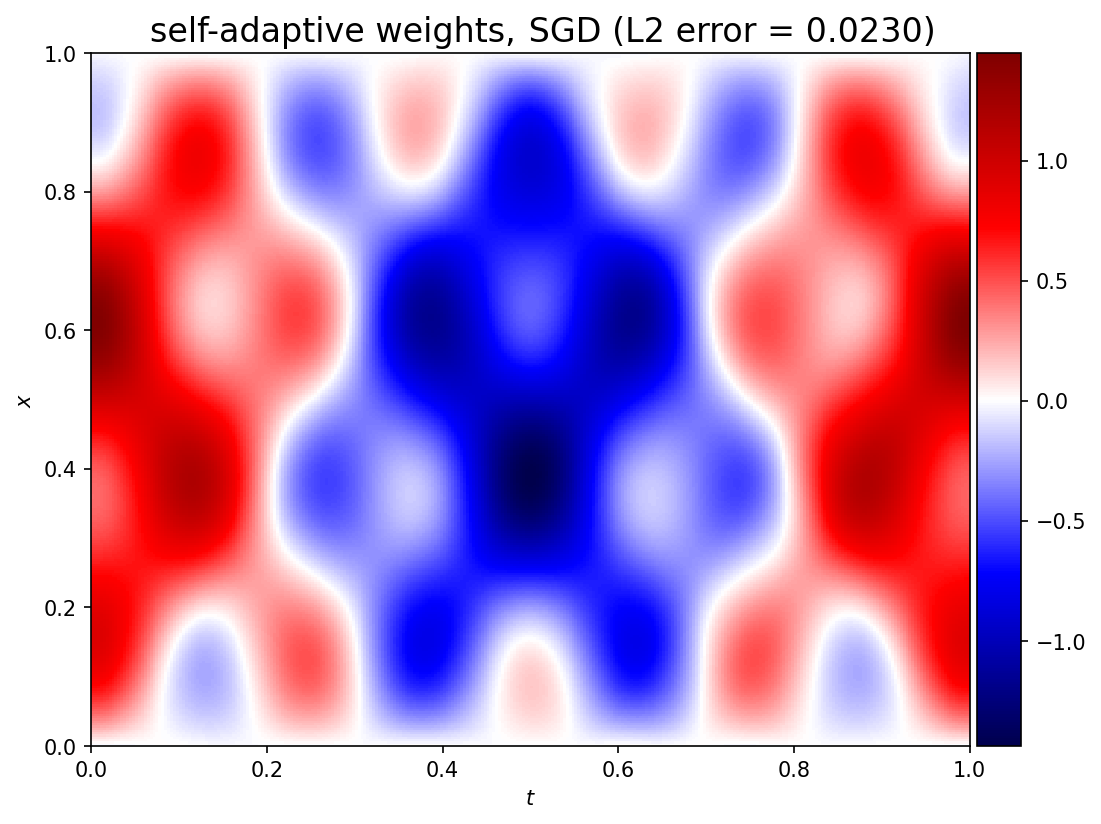}    
\end{tabular}
\vspace{-2.5ex}
\caption{\textit{Top:} Exact solution of the wave problem. \textit{Left:} Approximations obtained without SGD.  \textit{Right:} Approximations with SGD. From top to bottom: baseline, fixed weights, and self-adaptive weights.} 
\label{fig:wave_GP}
\end{figure}

\begin{figure}[hbt!]
    \centering
    \begin{tabular}{c}
    \includegraphics[width=.5\linewidth]{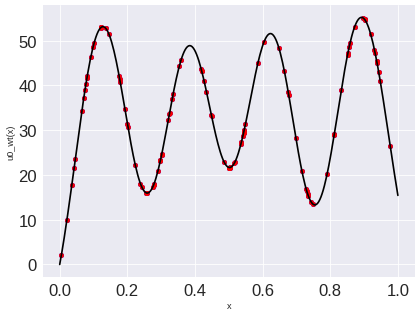} \\
    \includegraphics[width=.5\linewidth]{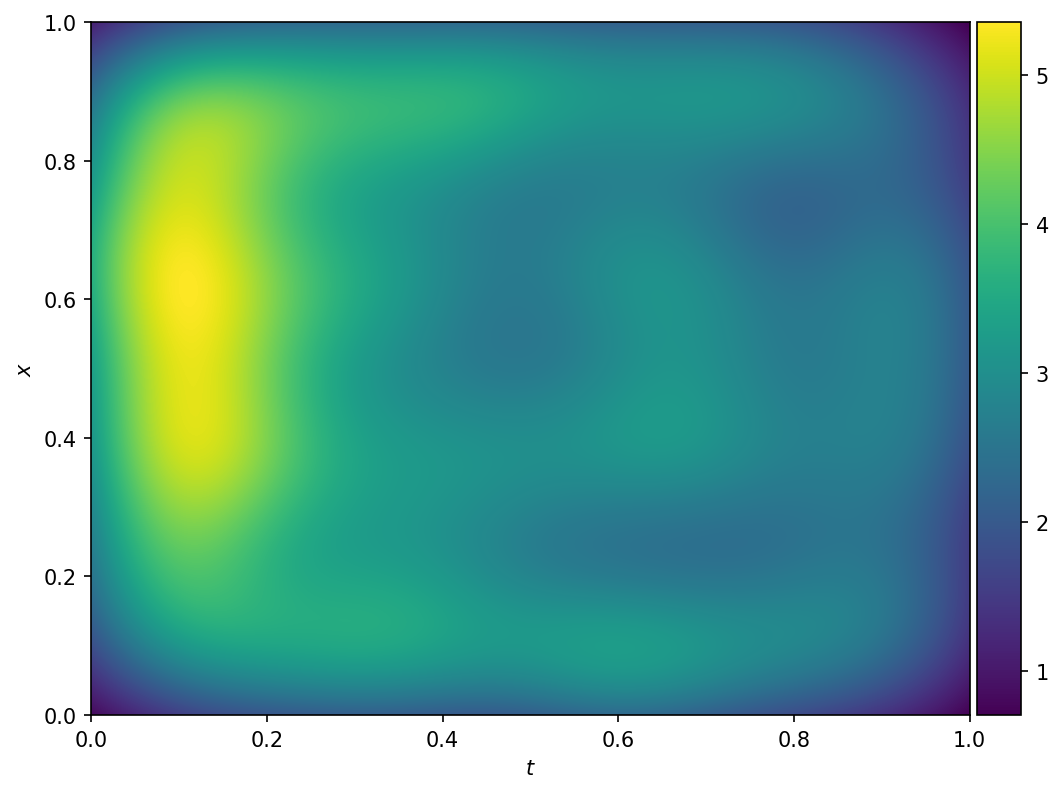}    
\end{tabular}
\vspace{-2.5ex}
\caption{Gaussian-Process maps of self-adaptive weights. \textit{Top:} 1D map of initial condition weights. The red dots indicate the values of the weights at the actual data locations. \textit{Bottom:} 2D map of PDE residue weights.} 
\label{fig:GP_wave_map}
\end{figure}

\begin{table}[!ht]
    \centering
\begin{small}
\begin{tabular}{ c|c|c|c|c } 
        \hline
            \multirow{2}{*}{PINN method} & \multicolumn{2}{c|}{No SGD} & \multicolumn{2}{c}{SGD}  \\
        \cline{2-5}
            & L2 error & time (sec) & L2 error & time (sec)\\
        \hline
            baseline & $0.3792 \pm  0.0162$ & 879.97 & $0.4513 \pm 0.0255$ & 1057.04\\
            fixed weights & $0.7296 \pm 0.1421$ & 850.75 & $0.2079 \pm 0.0624$ & 1012.06\\           self-adaptive & $0.8105 \pm 0.1591$ & 961.27 & $\mathbf{0.0295 \pm 0.0070}$ & 1207.85\\
         \hline
    \end{tabular}
\end{small}    
    \caption{Wave PDE results. The L2 error mean and standard deviation are based on 10 independent runs. The training time is an average over the 10 runs.}
    \label{table:SGD}
\end{table}

These results can perhaps be better appreciated in the plots in Figure~\ref{fig:wave_GP}. As an extra feature, the Gaussian Process predictor produces a {\em continuous} self-adaptive weight map. Figure~\ref{fig:GP_wave_map} displays the self-adaptive weight GP maps for the initial condition and residue points. We can observe in the 1D map that the self-adaptive weights become larger at the (high and low) peaks of the initial condition $u(x,0)$, which is where the curvature is maximum in magnitude; these are the most difficult regions to approximate with the neural network. In the 2D map, we can see that the self-adaptive weights are larger in the initial time, once again indicating the importance of approximating the solution early in time-evolution problems.

\section{Neural Tangent Kernel Analysis of Self-Adaptive PINNs}

\def\t{\tau}

In this section, we investigate the dynamics of SA-PINN training by studying its neural tangent kernel (NTK). We derive the expression for the NTK matrix for self-adpative PINNs and then use it to obtain a heuristic understanding of the effect of the self-adaptive weights on the dynamics of training in the limiting case of infinitely-wide PINNs. We examine the effect of the self-adaptive weights on the eigenvalues of the NTK matrix in the solution of a linear advection PDE.

First, note that (\ref{eq-gdw}) can be written as
\begin{equation}
   \frac{\bw^{k+1}-\bw^k}{\eta_k} \,=\,  - \nabla_{\bw}\mathcal{L}(\bw^k, \bl^k_r, \bl^k_b, \bl^k_{0})\,.
\end{equation}
In the limit as the learning rate $\eta_k$ tends to zero, the previous expression yields the {\em gradient flow} differential equation \cite{saxe2013exact}:
\begin{equation}
   \frac{d\bw(\t)}{d\t} \,=\, - \nabla_{\bw}\mathcal{L}(\bw(\t), \bl_r(\t), \bl_b(\t), \bl_0(\t))\,,
\label{eq-gradflow}
\end{equation}
where $\t\geq 0$ denotes the (continuous) training time. Notice that the usual gradient descent step corresponds to a forward Euler discretization of (\ref{eq-gradflow}). It follows that the properties of gradient descent optimization can be investigated by studying this differential equation.

Under this vanishing learning-rate limit, the {\em neural tangent kernel} (NTK) \cite{jacot2018neural} characterizes the training dynamics of the neural network, i.e., the  evolution of the output $u(\bx,t;\bw(\t))$ as a function of training time $\t$. In \cite{wang2020and}, the NTK for PINNs was derived and its properties were studied. Here we show how that a simple modification to their derivation produces the NTK for SA-PINNs.

For definiteness, consider the PDE problem:
\begin{align}
    & \mathcal{N}_{\bx}[ u(\bx)] \,=\, f(\bx,t)\,, \ \ \bx \in \Omega\,, \\
    & u(\bx) = g(\bx)\,, \ \
    \bx \in \Gamma \subseteq \partial{\Omega}\,.
\end{align}
For a time-evolution problem, $t$ becomes one of the components of $\bx$, and the set $\Gamma$ typically includes an initial condition at $t = 0$. More complex boundary conditions and sample data can be added to the analysis below in a straightforward way.

Given residue points $\{\bx_r^i\}_{i = 1}^{N_r}$ and boundary condition points  $\{\bx_b^i\}_{i = 1}^{N_b}$, let the response vectors be
\begin{align}
 {\bf u}_r(\t) &\,=\, [N_{x}[u(\bx^1_r;\bw(\t))],\ldots, N_{x}[u(\bx^{N_r}_r;\bw(\t))]]^T,\\ 
 {\bf u}_b(\t) &\,=\, [u(\bx^1_b;\bw(\t)),\ldots, u(\bx^{N_b}_r;\bw(\t))]^T.
\end{align}
Likewise, the data vectors are denoted by
\begin{align}
 {\bf v}_r &\,=\, [f(\bx^1_r),\ldots, f(\bx^{N_r}_r)]^T,\\ 
 {\bf v}_b &\,=\, [g(\bx^1_b),\ldots,g(\bx^{N_b}_b)]^T.
\end{align}
We write ${\bf u}_p(\t) = (u^1_p(\t),\ldots,u^{N_p}_p(\t))$ and ${\bf v}_p = (v^1_p,\ldots,v_p^{N_p})$ to identify the individual responses $u^i_p(\t)$ and data point $v^i_p$, for $p=r,b$. 

The loss function at training time $\t$ can be written similarly to (\ref{Lr})--(\ref{L0}):
\begin{equation}
\mathcal{L}(\bw(\t), \bl_r(\t), \bl_b(\t)) \,=\,  \frac{1}{2}\sum_{q=r,b} \sum_{j=1}^{N_q} m(\l^j_q(\t))\,|u^j_q(\t) - v^j_q|^2 
\end{equation} 
Hence, the gradient flow in (\ref{eq-gradflow}) becomes
\begin{align}
\frac{d{\bf w}}{d\t} & \,=\, - 
\sum_{q=r,b} \sum_{j=1}^{N_q} \nabla_{\bw} u_q^j(\t) m(\l^j_q(\t))\,(u^j_q(\t) - v^i_q) \\
&\,=\, - \sum_{q=r,b} {\bf J}_q^T(\t){\bf \Gamma}_q(\t)({\bf u}_q(\t) - {\bf v}_q)
\end{align}
where ${\bf J}_q(\tau)$ is the Jacobian of ${\bf u}_q(\t)$ with respect to $\bw$, for $q=r,b$, and ${\bf \Gamma}_q(\t)$ is a diagonal matrix of dimension $N_q \times N_q$ containing the self-adaptive mask values $m(\lambda_q^1(\t)),\ldots,m(\lambda_q^{N_q}(\t))$ in the diagonal, for $q = r,b$.


It follows that
\begin{align}
  &\frac{d {\bf u}_p(\t)}{d\t}  \,=\, {\bf J}_p(\t) \cdot \frac{d{\bf w}(\t)}{d\t} \,=\, - \sum_{q=r,b} {\bf J}_p(\t){\bf J}_q^T(\t){\bf \Gamma}_q(\t)({\bf u}_q(\t) - {\bf v}_q)\,,
\end{align}
for $p=r,b$. 

Now define
  ${\bf K}_{pq}(\t) = {\bf J}_p(\t){\bf J}_q^T(\t)$, 
  for $p,q = r,b$.
Notice that these are matrices of dimensions $N_p \times N_q$, with $i,j$ elements 
\begin{align}
  \left({\bf K}_{pq}\right)_{ij}(\t) \,=\, 
  \nabla_{\bw} u_p^i(\t)^T \cdot \nabla_{\bw} u_q^j(\t) \,=\, \sum_{w \in {\bf w}} \frac{d u_p^i(\t)}{dw} \cdot \frac{d u_q^j(\t)}{dw}\,.
\end{align}
It is clear from the definition that the matrices ${\bf K}_{pp}(\t)$ are symmetric and positive semi-definite, and that ${\bf K}_{pq}(\t) = {\bf K}_{qp}(\t)^T$, for $p,q = r,b$.

This allows us to collect the previous results in the following differential equation describing the evolution of the output of the SA-PINN in the vanishing learning-rate limit:
\begin{equation}
 \frac{d{\bf u}(\t)}{d\t}  \,=\, - {\bf K}(\t) \cdot ({\bf u}(\t) - {\bf v})\,,
\label{eq-NTK_system}
\end{equation} 
where 
\begin{equation}
  {\bf u}(\t) \,=\, \begin{bmatrix}
  {\bf u}_r(\t) \\
  {\bf u}_b(\t) 
 \end{bmatrix}, \quad\quad
  {\bf v} \,=\, \begin{bmatrix}
  {\bf v}_r \\
  {\bf v}_b
 \end{bmatrix}, 
\end{equation}
and
\begin{equation}
 {\bf K}(\t) \,=\, 
  \begin{bmatrix}
  {\bf K}_{rr}(\t){\bf \Gamma}_r(\t) & {\bf K}_{rb}(\t){\bf \Gamma}_b(\t) \\
  {\bf K}_{br}(\t){\bf \Gamma}_r(\t) & {\bf K}_{bb}(\t){\bf \Gamma}_b(\t) 
  \end{bmatrix}
\end{equation}
is the {\em empirical neural tangent kernel} matrix for the SA-PINN. (When all the mask values are 1, this reduces to the expression in Lemma~3.1 of \cite{wang2020and}.)


Next, we employ the previous result to perform a heuristic analysis of the self-adaptive weights through their effects in the gradient flow ODE system in (\ref{eq-NTK_system}). The analysis is based on the infinite-width limit of neural networks, when it can be shown that, under the vanishing learning rate regime, the NTK matrix converges to a constant deterministic value throughout training~\cite{jacot2018neural}. Under certain regularity conditions, it is shown in \cite{wang2020and} that this result still holds in the case of PINNs, i.e., the matrices ${\bf K}_{rr}(\tau)$, ${\bf K}_{rb}(\tau) = {\bf K}_{br}^T(\tau)$ and ${\bf K}_{bb}(\tau)$ are constant and equal to their respective values at initialization ($\tau = 0$) throughout training. In  \cite{wang2020and}, this was proved for PINNs with one hidden layer and linear PDEs, though the authors conjectured that this result also holds for multiple-layer PINNs and nonlinear PDEs.

We thus make the assumption that for a wide PINN under a small learning rate, the NTK matrix and self-adaptive weights change little during training, i.e.,
${\bf K}_{pq}(\t) \approx {\bf K}_{pq}$ and ${\bf \Gamma}_p(\tau) \approx {\bf \Gamma}_p$,
for $p,q = r,b$, $\t \geq 0$. In addition, we make the 
the simplifying approximation that the ODE system (\ref{eq-NTK_system}) can be decoupled, so that
\begin{equation}
 \frac{d{\bf u}_p(\t)}{d\t}  \,\approx\, - {\bf K}_{pp} {\bf \Gamma}_p \cdot ({\bf u}_p(\t) - {\bf v})\,, 
\label{eq-NTK_system2}
\end{equation} 
for $p = r,b$. (This approximation is also made,
implicitly, in Section~7.3 of~\cite{wang2020and}.) Some justification for the decoupling approximation comes from empirical evidence (not shown) that the matrix norms of the cross-terms ${\bf K}_{rb}{\bf \Gamma}_{b}$ and ${\bf K}_{br}{\bf \Gamma}_r$ are smaller than those of ${\bf K}_{rr}{\bf \Gamma}_r$ and ${\bf K}_{bb}{\bf \Gamma}_b$ and, in some cases, much smaller. This approximation allows us to gain a qualitative understanding of the importance of the residual and boundary loss components separately.

For $p = r,b$, matrix ${\bf K}_{pp}$ is real symmetric and positive semi-definite, and thus diagonalizable with nonnegative eigenvalues. However, matrix ${\bf K}_{pp}{\bf \Gamma}_p$ is not symmetric and not diagonalizable, in general. Fortunately, with  the extra minor assumption that ${\bf K}_{pp}$ is positive definite, and thus invertible, ${\bf K}_{pp}{\bf \Gamma}_p$ is diagonalizable. To see this, note that
\begin{equation}
  {\bf K}_{pp}^{-\frac{1}{2}}{\bf K}_{pp}{\bf \Gamma}_p {\bf K}_{pp}^{\frac{1}{2}} \,=\, {\bf K}_{pp}^{\frac{1}{2}}{\bf \Gamma}_p {\bf K}_{pp}^{\frac{1}{2}}\,. 
\end{equation}
But ${\bf K}_{pp}^{\frac{1}{2}}{\bf \Gamma}_p {\bf K}_{pp}^{\frac{1}{2}}$ is a product of symmetric matrices, and thus symmetric itself. Hence, ${\bf K}_{pp}{\bf \Gamma}_p$ is similar to a real symmetric matrix, and thus diagonalizable. Furthermore, it is fairly simple fact of matrix theory that if
$\gamma_p^1 \geq \cdots \geq \gamma_p^{N_p}$ and $\mu_p^1 \geq \cdots \geq \mu_p^{N_p}$ are the ordered eigenvalues of ${\bf K}_{pp}$ and ${\bf K}_{pp}{\bf \Gamma}_p$, respectively, and $\lambda_p^1, \geq \cdots \geq \lambda^{N_p}$ are the self-adaptive weights sorted by magnitude, then
\begin{align}
  \mu_p^1 & \:\leq\: m(\lambda_1)\,\gamma_p^1\,, \label{eq-mineigena}\\[0.5ex]
  \mu_p^n & \:\geq\: m(\lambda_n)\,\gamma_p^n\,. \label{eq-mineigenb}
\end{align}
In particular, (\ref{eq-mineigenb}) implies that all eigenvalues of ${\bf K}_{pp}{\bf \Gamma}_p$ are nonnegative (and positive, if ${\bf K}_{pp}$ is positive definite).

It follows that, under the assumption that ${\bf u}_p(0) \approx {\bf 0}$ (this can be achieved with proper initialization of the neural network weights), the solution of the ODE (\ref{eq-NTK_system2}) is given by
\begin{equation}
  {\bf u}_p(\t) \\
  \,=\, 
 ({\bf I} - e^{-{\bf K}_{pp}{\bf \Gamma}_p t})\cdot  
  {\bf v}_p\,,
\end{equation}
which can be rewritten as
\begin{equation}
  {\bf u}_p(\t) \!-\! {\bf v}_p 
  \,=\, 
 - {\bf Q}^Te^{-{\bf M}t}{\bf Q}\cdot  
  {\bf v}_p\,,
\end{equation}
that is 
\begin{equation}
  {\bf Q} \cdot
  ({\bf u}_p(\t) \!-\! {\bf v}_p)
  \,=\, 
 - e^{-{\bf M}t}{\bf Q}\cdot {\bf v}_p\,,
\end{equation}
where ${\bf Q}$ is the matrix of eigenvectors and ${\bf M}$ is the diagonal matrix of eigenvalues $\mu_p^1,\ldots,\mu_p^{N_p}$ of ${\bf K}_{pp}{\bf \Gamma}_p$, for $p = r,b$. This implies that
the training error $u_p^i(\t) - v_p^i$ decreases at a rate $e^{-mu_p^i}$ rate. Large variation among the eigenvalues $\mu_p^1,\ldots,\mu_p^{N_p}$, both across the different loss terms $p=r,b$ {\em and} the different data points in each loss term, will potentially lead to training imbalances and loss of convergence. 

The standard weighted loss function
in (\ref{eq:loss_weighted}) corresponds to the case when all the self-adaptive weights for each loss component are equal, with ${\bf \Gamma}_p = \lambda_p I$, in which case the eigenvalues of the NTK matrix are simply scaled by $\lambda_p$: $\mu_p^i = \lambda_p^i \gamma_p^i$, for $i=1,\ldots,N_p$. On the other hand, the transformation effected on the eigenvalues of the NTK matrix by the self-adaptive weights is nonlinear. In general, little can be said about it other than the transformed eigenvalues are in the interval determined by $m(\lambda_1)\gamma_p^1$ and $m(\lambda_n)\gamma_p^n$, as stated in (\ref{eq-mineigena})--(\ref{eq-mineigenb}). The simple linear scaling introduced by traditional weighting can certainly help reduce the imbalance among the various terms, but it is less flexible than the transformation introduced by the pointwise self-adaptive weights, which can also change the {\em shape} of the eigenvalue distribution. 

Next, we illustrate this analysis with the classical univariate advection PDE:~\cite{leveque2002finite}:
  \begin{align}
  & q_t(x,t) + \bar{u}q_x(x,t) \,=\, 0\,, \ \  x \in [0,L]\,, \ t \in [0,T]\,,\\
    & u(0,t) = u(L,t) = 0, \ t \in [0,T]\,,\\
    & u(x,0) = g(x)\,, \ x \in [0,L]\,. 
\end{align}
where $q(x,t)$ is for example the concentration of a tracer being transported in a fluid in a tube of length $L$, where $\bar{u}>0$ is the fluid constant velocity. For simplicity, it is assumed that $g(x) = 0$ outside an interval in $[0,L]$, and that $T$ is short enough that the Dirichlet boundary condition is satisfied. (This could be changed at the expense of more complex boundary conditions.) In this  scenario, the problem has a simple solution:
\begin{equation}
     q(x,t) \,=\, g(x-\bar{u}t)\,, \ \ x \in [0,L]\,, \ t \in [0,T]\,, 
\end{equation} 
i.e., the initial concentration profile is simply translated to the right at constant speed $\bar{u}$. Here, we adopt a fairly complex initial condition, containing several discontinuities, which makes the problem rather difficult to solve with the baseline PINN --- or indeed numerical methods in general~\cite{leveque2002finite}.

\begin{figure}[hbt!]
    \centering
    \includegraphics[width=1\linewidth]{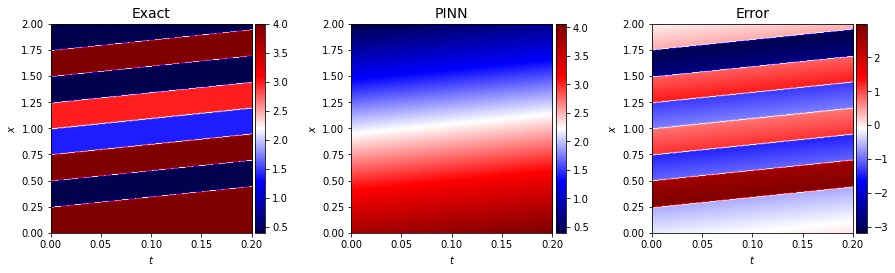}\\
    \includegraphics[width=1\linewidth]{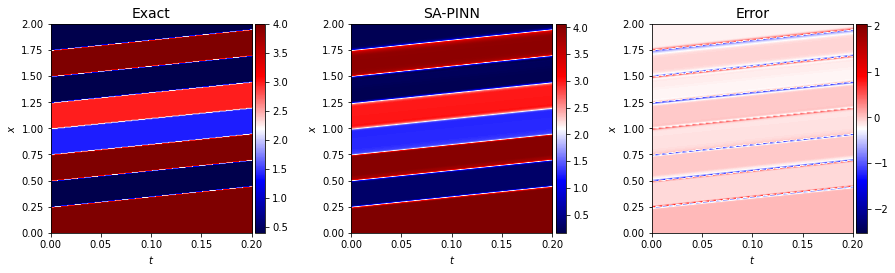}
    \caption{\textit{Top:} Plot of the approximation $u(x,t)$ via the baseline PINN, showing the exact solution vs predicted solution vs absolute error. \textit{Bottom:} The SA-PINN results, L2 error decreases by an order of magnitude and the SA-PINN closely captures the exact solution. 
}
    \label{fig:trained_advection}
\end{figure}

\begin{figure}[hbt!]
    \centering
    \includegraphics[width=1\linewidth]{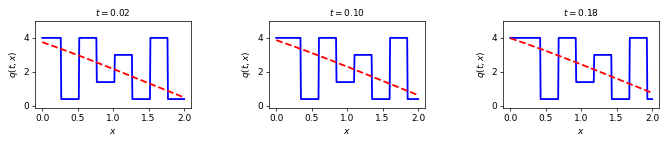}\\
    \includegraphics[width=1\linewidth]{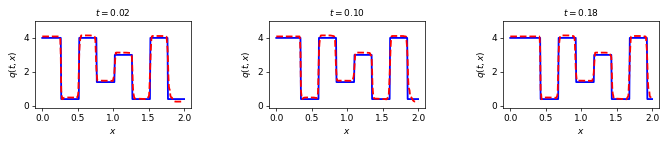}
    \caption{\textit{Top:} Plot of the approximation $u(x,t)$ via the baseline PINN, showing cross sections of the spatial domain at $t=0.02, 0.10, 0.18$ \textit{Bottom:} The SA-PINN results at the same time steps, with the same number of epochs (10k Adam) and all other parameters held constant. 
}
    \label{fig:trained_advection_cross}
\end{figure}

\begin{figure}[hbt!]
    \centering
    \includegraphics[width=.5\linewidth]{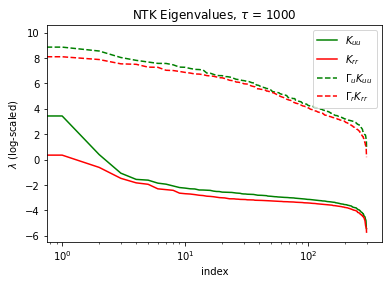} \\
    \includegraphics[width=.5\linewidth]{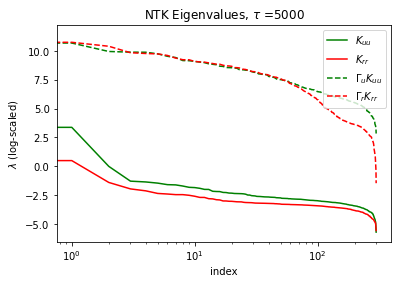} \\
    \includegraphics[width=.5\linewidth]{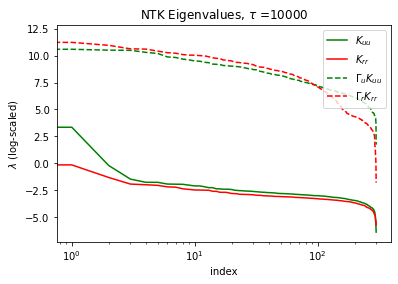}
    \caption{NTK eigenvalues of the baseline PINN (solid) vs. the SA-PINN (dashed) for $\tau=$1000, 5000, and 10000 training iterations. It can be observed that the SA-PINN accurately matches the magnitudes of the NTK eigenvalues between terms of the loss function, in this case the initial condition $K_{uu}$ and the residual loss $K_{rr}$}
    \label{fig:trained_advection_eigs}
\end{figure}

The results presented in figures~\ref{fig:trained_advection},~\ref{fig:trained_advection_cross}, and~\ref{fig:trained_advection_eigs} are generated with a neural network architecture of [2, 400, 400, 400, 400, 1], trained for 10k Adam iterations with a neural network weight learning rate of 0.001 (hence, a wide PINN with a smal learning rate, as required by the theory). The learning rate for all self-adaptive weight was set at 0.1. Glorot Normal initialization was utilized, and all training was completed in Tensorflow on a single V100 GPU with an average training time of 7 seconds for 10k iterations. At the end of 10k training iterations, the baseline PINN failed to grasp even the rough structure of the solution, while the SA-PINN was able to approximate the solution within ~5\% L2 error. (More accurate results could have obtained by using more training epochs and a decreasing learning rate schedule.)

An analysis of NTK eigenvalues similar to that performed in~\cite{wang2020and} is demonstrated in figure~\ref{fig:trained_advection_eigs}. (In this example, the boundary condition weights were fixed and equal to 1.0, and we disregarded this component in the analysis.) We can see that the eigenvalues become closely matched in scale between $K_{uu}$ and $K_{rr}$, removing the imbalance between these two loss components and enabling convergence to the solution (here we are only looking at the initial and . Importantly, the shape of the eigenvalue distribution is also nicely equalized, as opposed to simply being scaled up as would be the case with traditional weighting of the entire loss component.

\section{Conclusion}

In this paper, we introduced Self-Adaptive Physics-Informed Neural Networks, a novel class of physics-constrained neural networks. This approach uses a similar conceptual framework as soft self-attention mechanisms in computer vision, in that the network identifies which inputs are most important to its own training. It was shown that training of the SA-PINN is formally equivalent to solving a PDE-constrained optimization problem using penalty-based method, though in a way where the monotonically-nondecreasing penalty coefficients are trainable. Experimental results with several linear and nonlinear PDE benchmarks indicate that SA-PINNs produce more accurate solutions than other state-of-the-art PINN algorithms. It was seen that SA-PINNs can employ stochastic gradient training, through continuous Gaussian-process interpolated self-adaptive maps, which allows the solution of a difficult wave PDE. These experimental results were complemented by a theoretical analysis based on the Neural Tangent Kernel for SA-PINNs.

We believe that SA-PINNs open up new possibilities for the use of deep neural networks in forward and inverse modeling in engineering and science. However, there is much that is not known yet about this class of algorithms. While the empirical results observed here show significant promise, and despite the fact that an initial analysis based on the NTK is provided, more theoretical justification is desirable, which will be part of future work. In addition, the use of standard off-the-shelf optimization algorithms for training deep neural networks, such as Adam, may not be appropriate, since those algorithms were mostly developed for traditional deep learning applications; obtaining optimization algorithms specifically tailored to Self-Adaptive PINNs, and indeed PINNs in general, in an open problem. Finally, the relationship between Self-Adaptive PINNs and constrained-optimization problems is likely a fruitful topic of future study.  

\section*{Acknowledgments}

The authors would like to acknowledge the support of the D$^3$EM program funded through NSF Award DGE-1545403. The authors would further like to thank the US Army CCDC Army Research Lab for their generous support and affiliation. 



\newpage

\bibliographystyle{unsrt} 
\bibliography{ArXiv_v5}

\end{document}